\documentclass[sigconf]{acmart}
\AtBeginDocument{%
  \providecommand\BibTeX{{%
    \normalfont B\kern-0.5em{\scshape i\kern-0.25em b}\kern-0.8em\TeX}}}

\setcopyright{acmcopyright}
\copyrightyear{2021}
\acmYear{2021}
\usepackage{microtype}
\usepackage{graphicx}
\usepackage{caption}
\usepackage{subcaption}
\usepackage{booktabs} % for professional tables
\usepackage{tabularx}
\usepackage{multicol}
\usepackage{empheq}
\usepackage{amsmath}
\usepackage{todonotes}
\usepackage{xspace} 
\usepackage{multirow}
\usepackage[ruled,norelsize,linesnumbered]{algorithm2e}
\usepackage{algorithmic}
\usepackage{tikz}
\usepackage[
  separate-uncertainty = true,
  multi-part-units = repeat
]{siunitx}

\newcommand{\approach}{Elixir}
\newcommand{\etc}{\emph{etc.}\xspace}

\newcommand{\ie}{\emph{i.e.,}\xspace}
\newcommand{\eg}{\emph{e.g.,}\xspace}
\newcommand{\etal}{\emph{et al.}\xspace}

\newcommand{\cut}[1]{}

\usepackage{hyperref}

\newcommand{\figref}[1]{Figure \ref{#1}}
\newcommand{\secref}[1]{Section \ref{#1}}
\newcommand{\tabref}[1]{Table \ref{#1}}

\renewcommand\footnotetextcopyrightpermission[1]{} % removes footnote with conference information in first column

\title{\approach: A system to enhance data quality for multiple analytics on a video stream}
\author{Sibendu Paul}
\affiliation{%
  \institution{Purdue University}
  \city{West Lafayette}
  \country{USA}
}
\author{Kunal Rao}
\affiliation{%
  \institution{NEC Laboratories America, Inc.}
   \city{New Jersey}
  \country{USA}
}
\author{Giuseppe Coviello}
\affiliation{%
  \institution{NEC Laboratories America, Inc.}
   \city{New Jersey}
  \country{USA}
}
\author{Murugan Sankaradas}
\affiliation{%
  \institution{NEC Laboratories America, Inc.}
   \city{New Jersey}
  \country{USA}
}
\author{Oliver Po}
\affiliation{%
  \institution{NEC Laboratories America, Inc.}
   \city{San Jose}
  \country{USA}
}
\author{Y. Charlie Hu}
\affiliation{%
  \institution{Purdue University}
  \city{West Lafayette}
  \country{USA}
}
\author{Srimat Chakradhar}
\affiliation{%
  \institution{NEC Laboratories America, Inc.}
   \city{New Jersey}
  \country{USA}
}

\begin{document}
\begin{abstract}
%Video cameras are now-a-days ubiquitously deployed 

IoT sensors, especially video cameras, are ubiquitously deployed 
around the world to perform a variety of computer vision tasks in
several verticals including retail, healthcare, safety and security,
transportation, 
% entertainment, 
manufacturing, etc. To amortize
their high deployment effort and cost,
it is desirable 
% to get the most out of every camera video feed. This motivates 
to perform multiple video analytics tasks, which we refer to as
Analytical Units (AUs), off the video feed coming out of every camera.
%
%   As an example, AUs like crowd/people counting, facial recognition, age
%   and gender recognition, suspicious activity recognition, intrusion
%  detection, car counting, license plate recognition, etc. are performed
%  on outdoor cameras or surveillance cameras deployed to monitor road
% intersections, enterprise buildings, retail stores, etc. 
As AUs typically use deep learning-based AI/ML models, their
performance depend on the quality of the input video, and
recent work has shown that dynamically adjusting the camera setting
exposed by popular network cameras can help improve the quality of the
video feed and hence the AU accuray, in a single AU setting.

In this paper, we first show that 
in a {\em   multi-AU setting},
changing the camera setting has
disproportionate impact on different AUs performance.
In particular, the optimal setting for one AU may
severely degrade the performance for another AU, and further the
impact on different AUs varies as the environmental condition changes.
%i.e. one camera setting is good for one AU, while some other completely different camera setting is good for another AU on the same video camera. 
%  Also, the camera settings that give optimal performance for any given
%  AU are not the same always i.e. they change as the environment and
%  video content changes, which is shown in ~\cite{paul2021camtuner} for
%  \textit{single} AU. 
%   Therefore, it is very challenging to dynamically
%  %  find optimal camera settings that improves accuracy for
%  \textit{multiple} AUs simultaneously.
%Therefore, it is a non-trivial challenge to dynamically adjust camera settings such that all the AUs deliver good performance.

We then present \approach, a system to enhance the video stream quality for
\textit{multiple} analytics on a video stream.
%  by dynamically changing the video camera setting. 
\approach\ leverages Multi-Objective
Reinforcement Learning (MORL), where the RL agent caters to the
objectives from different AUs
%  deployed on a single video camera, 
and adjusts the camera setting to 
simultaneously enhance the performance of all AUs.
%    while taking corrective actions
%  related to camera settings adjustment. 
To define the multiple objectives in MORL, we develop new
AU-specific quality estimator values for each individual AU.
%  and the goal for the RL agent is to maximize the value of a function, which
% incorporates the quality estimate values from the different AUs. 
%
We evaluate \approach\ through real-world experiments
on a testbed with three cameras
deployed next to each other (overlooking a large enterprise parking lot)
running Elixir and two baseline approaches, respectively.
%  one with the default camera setting, the other using time-sharing technique
%  for camera settings adjustment, and the last one with \approach.
% Our results show that the ORL-based
% enhance data quality, such that \comment{
Elixir correctly detects 7.1\% (22,068) and 5.0\% (15,731) more cars, 94\%
  (551) and 72\% (478) more faces, and 670.4\% (4975) and 158.6\%
  (3507) more persons than the default-setting and
  time-sharing approaches, respectively. It also detects 115 license
  plates, far more than the time-sharing approach (7) and the default
  setting (0).  

\end{abstract}

\maketitle
\vspace{-0.1in}
\section{Introduction}
%put references for the growing video analytics market
%Due to the spurring demand for automation and the impetus to augment smart technologies, the global smart sensor market size is expected to grow from \$36 billion in 2020 up to \$90 billion by 2027 at a CAGR of 13\% \textcolor{blue}{[add citation]}.

Due to the spurring demand for automation and the impetus to make
everything ``smart", IoT sensors are increasingly being deployed and
there has been rapid technological advancement to utilize raw sensor
data and generate actionable insights from them. The global smart
sensor market size is expected to grow from \$37 billion in 2020 to
\$90 billion by 2027 at a CAGR of
14\%~\cite{digital-market-research}. Among these IoT sensors, video
cameras are very popular, since they act as our ``eyes" to the world
and many important video analytics tasks, which we refer to as Analytical
Units (AUs), are performed on the raw visual data feed from video
cameras.

\begin{figure*}[tp]
    \centering
        \includegraphics[width=0.89\textwidth]{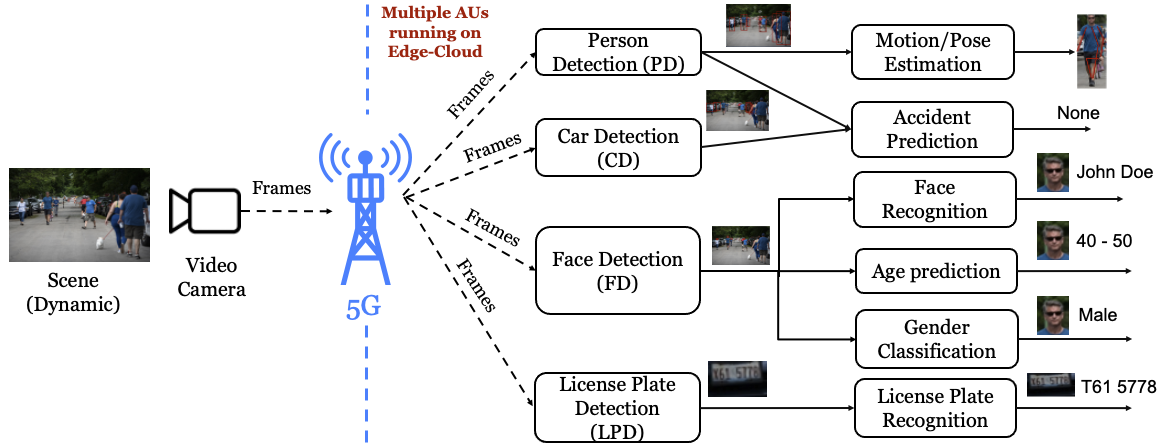}
\caption{Multiple AUs running on a single video stream.}
\label{fig:mulAU_Retail}
\vspace{-0.1in}
\end{figure*}

\begin{figure*}
    \centering
        \includegraphics[width=0.86\textwidth]{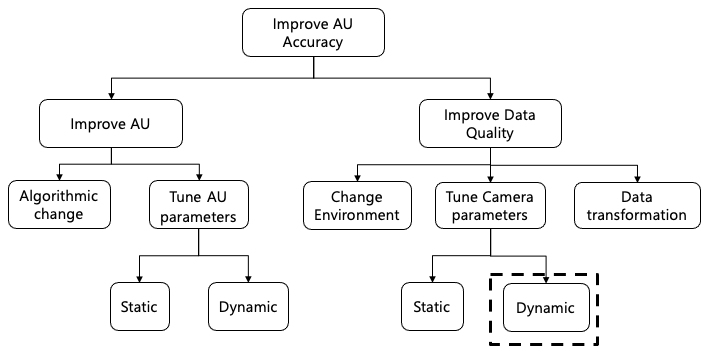}
\caption{A taxonomy of possible techniques to improve AU accuracy in video analytics.}
\label{fig:positioning}
\vspace{-0.1in}
\end{figure*}

Deploying cameras requires enormous effort and investment, and
thus there is often expectation to get the most out of the
deployed cameras, by running multiple AUs on each video stream
to generate multiple, different insights. 
%in a mall/retail store, 
For example, as shown in ~\figref{fig:mulAU_Retail}, from a single
video camera feed, AUs like person detection, motion/pose estimation,
face detection, face recognition (to identify VIP/person of interest),
age prediction, gender classification, license plate detection,
license plate recognition, accident prediction, etc. can be performed.
For a camera deployed in a store, further analytics such as the number of
unique visitors, flow of people inside the store, congregation of
people, time spent by visitors in different sections, etc. can be
performed.
%the number of people that enter/exit, the demographics of people, congregation of people, identification of VIP or person of interest, flow of people, time spent in different sections, etc. is captured using multiple different AUs. 
All such information is very useful for store safety, store analytics, 
and improvement of conversion rate. As another example, for video feed off a camera
deployed at a road intersection, multiple AUs like vehicle detection,
license plate detection and recognition, pedestrian detection, vehicle
speed detection, red light crossing detection, accident prediction,
etc. can be performed.
There are many more scenarios where a camera is deployed in an enterprise building,
stadium, bank, or factory, etc., 
and multiple different AUs are run at the same time to generate
insights. Sometimes high level AUs are dependent on multiple low level
AUs to generate insights and in such scenarios we need joint
optimization for multiple AUs. For example, 
for an accident prediction AU, detection of pedestrians, cars, their speed of movement, etc. all have to work well and these tasks serve as building blocks for the accident prediction AU.
%for an accident prediction AU, detection of pedestrians, cars, their speed of movement, \etc all have to happen accurately for the accident prediction AU to work well.

AUs typically use deep-learning based AI/ML models, which act as our
``brains" to process the visual data seen through the ``eyes" of video
cameras. These models are developed by building deep neural networks
(DNNs) and training these networks with thousands of images. This
process leads to some ``trained" models, that are capable of
performing specific computer vision tasks, e.g., object detection. As
such, it is very important for these models to be highly accurate for
the generated insights to be useful.
%Now, as shown in ~\figref{fig:positioning}, there are several possible dimensions to improve the accuracy of these AI/ML models, which are used within AUs. 

There are two general approaches towards improving the accuracy of an AU
in a video analytics system, as shown~\figref{fig:positioning}:
improving the AU design and improving the quality of video stream.
In the first approach,
one can improve the accuracy of the AU by 
making algorithmic changes in the underlying DNN model. 
This, however, is a very expensive and slow solution, because it is not
practical to keep training the model with new algorithms and new
datasets on costly GPU resources. 
Another way to improve the AU is by
tuning AU parameters, which can be done either at the start,
i.e., statically, or during operation, i.e., dynamically. 
One example of AU parameters is the minimum number of pixels
between eyes, which is used for ``face detection" AU. This
configurable parameter can be tuned to determine faces in the
scene. However, depending on the environmental conditions around the camera
(lighting conditions), the content in the scene (number of faces,
speed of movement of people), the settings of the camera (brightness,
contrast, color, sharpness, exposure, shutter speed), etc., the quality
of image obtained from the camera can vary. As a result, adjusting the configuration
value for ``minimum pixels between the eyes" after the image
acquisition from the camera may not be able to improve the AU
accuracy if the video frame itself is not of good quality.

%This assumes that the input visual data feed cannot be altered and therefore fall back on configuring the AU parameters. However, modern cameras expose several camera settings, which can be adjusted in order to alter the quality of video feed coming out of the camera, thereby avoiding the need to make configuration changes to the AU itself.

The second general approach to improving AU accuracy is to improve the quality of
data that is fed to the AU. 
Intuively, if the AU receives ``good'' quality input data, e.g., similar to
what it has seen during ``training'', the chances of the AU
producing accurate results is high. There are three different ways in
which we can improve the quality of input data to the AU, namely, 
(1) improving the environment around the camera, e.g., by adding additional
light, such that the camera is able to get a better capture of the
scene,
(2) applying data transformation on individual
frames of the video feed captured by the camera, and
(3) tuning camera parameters in order to help the camera capture
good quality video.
Changing the environment is not practical in different scenarios and
applying data transformation is typically too late because
the camera has already captured the scene. 
Therefore, in this paper, we focus on tuning camera
parameters. A similar approach has been studied in~\cite{paul2021camtuner},
which, however, has only considered single-AU scenarios.
%This camera parameter tuning can be done statically, one-time, but it is shown in ~\cite{paul2021camtuner} that such one-time tuning is not optimal as the environment and video content changes, and the tuning needs to be done dynamically. 
%In this paper, we take similar approach, but 
Unlike~\cite{paul2021camtuner}, in this paper, we consider
\textit{multiple} AUs at the same time and study how to dynamically tune
the camera setting for multiple AUs.
%  is done simultaneously, as shown in ~\figref{fig:positioning}.

Tuning camera parameters for a single AU itself is already 
challenging because the optimal camera parameters for an AU keep
changing as the environmental condition and the video
content change over time, and 
% it is very difficult to find the optimal 
the camera comes with thousands of possible camera settings even for a
handful of parameters, as shown in~\cite{paul2021camtuner}. 
When there are multiple different AUs running at the same time on a
single camera feed, the problem of dynamic camera settings adjustment
becomes even more challenging. The key reason is that the optimal camera
settings for one AU may degrade performance for another AU and vice
versa, as shown in our experiments.
In particular, each AU has some optimal camera setting but when
multiple AUs all work off the same camera feed, which can assume only
one physical camera setting at any given point in time, the camera
setting that is ``good" for all AUs at the same time is extremely
difficult to identify. In this paper, we tackle this challenging
problem and present \approach, a system that dynamically finds the
optimal camera setting that enhances input video quality such that the
performance of multiple different AUs, which are running off the
camera feed, are improved simultaneously.

%Thus, we incline towards dynamic camera parameter tuning in order to improve AU accuracy, while it is in operation.

%, all of them deliver ``good" performance to the extent possible.

\approach\ is an online system, meaning it dynamically tunes camera
settings while the AUs are in operation, i.e., in reaction to 
environmental condition and scene change.
Making such dynamic adjustment faces a fundamental design challenge
-- how to estimate the quality of the current camera setting, \ie, the
accuracy of the AUs, as it is not possible to measure AU
accuracy online using traditional methods, which require ground truth
information.
% (since it is not practical for humans to constantly provide ground truth information). 
To address this challenge, we develop and use new AU-specific quality
estimators, which analyze a frame and estimate the quality of the
frame for the specific AU accuracy. The higher the quality, the more likely
that the AU will have high accuracy on the frame. 
With the set of AU-specific quality estimators,
one for each AU, the goal of \approach\ is 
to dynamically identify the camera settings that will lead to
high values on the AU-specific quality estimates. To do so,
\approach\ leverages Multi-Objective Reinforcement Learning (MORL) in
which the multiple objectives correspond to the different AU-specific
quality estimates (one per AU)
and the goal for the RL agent is 
to maximize the combined value of 
an aggregate objective function that incorporates quality estimate values from all the AUs
running on the single video stream.

In summary, the key contributions of this paper are:
%\vspace{-0.05in}
\begin{itemize}
    %\item We show that accuracy of video analytics tasks, which we refer to as Analytical Units (AUs), can be improved by enhancing quality of input data by dynamically adjusting camera settings, but when multiple AUs run on a single video camera feed, then the optimal camera settings for one AU at any given point in time may degrade accuracy of another AU. \textcolor{blue}{rephrase a little bit}
\item We show that when \textit{multiple} AUs run on a single
      video camera feed, the optimal camera setting for one AU
      degrades accuracy of another AU, \ie they conflict with each       other.
% but there exists a common optimal setting that works well
%       for multiple AUs running off of a single video stream.
\item We present \approach, a system that leverages Multi-Objective
      Reinforcement Learning (MORL) to enhance the quality of
      input camera feed to AUs by dynamically finding the optimal camera setting
      by training the RL agent to maximize the aggregate quality estimate for all AUs.
%       which incorporates quality estimate values from all AUs running
%       on a single stream, thereby ensuring that the camera settings
%      are such that accuracy of all AUs is improved simultaneously.
    \item We develop new AU-specific quality estimators for multiple
      different AUs, including face detection, person detection, car
      detection and license plate detection, and show that these can
      be used as proxies for estimating AU accuracy in the absence of
      ground truth, which is not practical to obtain during real-time
      operation of the AUs.
    \item We perform real-world experiments with three cameras running
      multiple AUs on each cameras at the same time,
      wherein the first camera uses the default camera settings, the second one
      uses time-sharing technique to favor one particular AU during
      its time share, and the third uses \approach. Our results show that
      when compared to using the default camera settings and the
      time-sharing approach, Elixir correctly detects 7.1\% (22,068)
      and 5.0\% (15,731) more cars, 94\% (551) and 72\% (478) more
      faces, and 670.4\% (4975) and 158.6\% (3507) more persons,
% than       the default and time-sharing approaches, 
respectively. It also
      detects 115 license plates, far more than the time-sharing
      approach (7) and the default setting (0).
    %on average, \approach\ is able to improve accuracy of four different AUs i.e. person detection, face detection, car detection and license plate detection running at the same time on a single video stream by \textcolor{blue}{x\%} when compared with the closest alternative approach.
\end{itemize}

The rest of the paper is organized as follows. 
~\secref{related} discusses related work.
~\secref{sec:motivation} motivates the need for \approach.
Next,  ~\secref{section:challenge} 
discusses the various challenges encountered in designing \approach\ and
our approach to overcome those challenges. \approach's overall system
design is shown in ~\secref{sec:design} and results are presented in
~\secref{section:eval}. Finally, we discuss further possible improvements
using digital transformation in~\secref{sec:further_improvement} and
conclude in ~\secref{sec:conclusion}.

\section{Related Work}
\label{related}
Applying multiple AUs on a single camera video stream is explored in MultiSense~\cite{sharma2011multisense} and Panoptes~\cite{jain2017panoptes}, but they consider AUs which require different views from the camera. To support these AUs at the same time, they propose to use PTZ cameras, which can be remotely controlled to change the view of the camera as per the need of the AU. They multiplex multiple AUs through time-sharing approach and via view virtualization. This way, multiple AUs can run on a single video stream. However, in this paper, we consider multiple AUs that run together on the \textit{same} view of the camera and there is no need to control and change the view of the camera. In such a setup, we show that each AU has an optimal camera setting to achieve highest accuracy for the AU, but these optimal camera settings conflict with each other i.e. optimal camera setting for one AU degrades accuracy for another AU. Such conflict across multiple AUs is not considered in MultiSense~\cite{sharma2011multisense} and Panoptes~\cite{jain2017panoptes}.

Along with multiple AUs running off of a single camera stream, there are also several works which focus on single AU running off of multiple cameras.
%there are also multiple works on multi-camera surveillance networks that coordinate among each others to support only one analytics application. 
Most recent work, VeTrac~\cite{tong2021large} employs widely deployed traffic cameras as a sensing network for vehicle tracking at large scale. ~\cite{dao2017accurate, qureshi2009planning} propose to automatically capture high quality surveillance video through distributing targets to different co-located cameras. Yao \etal~\cite{yao2010adaptive} 
%consider limited camera computational capacity while 
adaptively assign several PTZ cameras based on the moving objects in a multi-camera, multi-target surveillance system. In this paper, we do not consider such multi-camera scenario, but focus on single camera, multiple AUs running off of this single camera.

Changes in environmental conditions have direct impact on the accuracy of the AU as shown in ~\cite{paul2021camtuner}. To alleviate the degradation in AU accuracy due to environmental changes, Jang \etal~\cite{jang2018application} propose IoT camera virtualization and support multiple AUs, each using different AI/ML model, which is appropriate for the specific environment. These multiple AUs, however, are to generate similar insight and only one of them runs at any given point in time. In contrast, in this paper, we consider multiple AUs, each generating different insights and they all run simultaneously at the same time.

To improve accuracy for single AU, few recent works, \eg AMS~\cite{khani2021real},
 Focus~\cite{222587}, Ekya~\cite{padmanabhantowards}, and NoScope~\cite{10.14778/3137628.3137664}, studied dynamic AU model parameters tuning based on captured video content.
%, which is shown along the left branch in ~\figref{fig:positioning}. 
Although it is a plausible approach to extend this method for multiple AUs, but the periodic model training on expensive GPU resources and slow reaction to dynamic changes, makes this approach non-feasible for practical deployment.

Several works \eg  Chameleon~\cite{jiang2018chameleon}, Videostorm~\cite{201465}, and AWStream~\cite{zhang2018awstream} adapt camera settings like frame resolution and frame per second (fps) while processing single AU. Their goal is to make it efficient to run video analytics tasks, while the goal for \approach\ is to improve AU accuracy for multiple AUs running off of a single video stream. Thus, although we could extend this work for multiple AUs, since the goal is different, we chose not to go with these camera settings, which make sense to change from efficiency point of view but not from improving AU accuracy point of view. 

%We could extend this to multiple AUs, but different AUs may not be amenable to changes in these settings. Therefore, in this paper, we focus on other camera settings, which would work for different AUs.

The most recent work, CamTuner~\cite{paul2021camtuner} dynamically tunes the four camera settings considered in this paper, in response to changes in environmental conditions in order to improve the accuracy of a video analytics pipeline. However, CamTuner focuses on improving accuracy of a single AU running on a camera stream. In practice, multiple AUs typically run off of a single camera stream (\eg FD, PD, CD, LPD, \etc as shown in~\figref{fig:mulAU_Retail}), and this multiple AUs scenario is not addressed by CamTuner. Similar to CamTuner, in this paper, we focus on the four camera settings i.e. brightness, contrast, color and sharpness, but the changes done to these settings in CamTuner is with respect to a single AU. In contrast, \approach\ dynamically changes these settings in order to cater to multiple AUs. In this paper, we use AU-specific quality estimator, which was introduced in CamTuner, but we develop new one's for PD, FD, CD and LPD AUs. In CamTuner, since there is only a single AU, there is no question of conflict across AUs, while in \approach, camera settings for one AU conflicts with another AU. For instance, an increase in brightness of the entire capture might be beneficial for coarse-grained person or car detection but it can hurt fine-grained face detection and facial recognition, as the face area get overexposed. Thus, challenges solved by \approach\ are different from the one's solved by CamTuner. To the best of our knowledge, this is the first work that addresses the challenge of improving accuracy for multiple AUs running off of a single camera stream by dynamically tuning camera settings, which enhances data quality such that accuracy of multiple AUs is improved simultaneously.

\vspace{-0.1in}
\section{Motivation}
\label{sec:motivation}

%In this section, we analyze the impact of using best camera parameter setting for one AU on the accuracy of insights derived by another AU when multiple AUs are deriving insights from the same video stream. Specifically, we consider video analytics applications in Retail Store and Intelligent Transportation Systems. 

%In this section, we initially show that for a given AU, the optimal camera setting changes depending on the environmental conditions i.e. different camera settings are optimal for an AU for different environmental conditions. Then, we 

In this section, we first show that different AUs benefit from
different camera settings, i.e., the optimal camera setting for
different AUs differ at any given point in time. Then, we
systematically study the impact of optimal camera settings for
different AUs and show optimal camera settings conflict, i.e., the optimal
camera setting for one AU degrades performance of another AU.

%In this section, we initially show that different AUs benefit from different camera settings i.e. the optimal camera setting for different AUs is not the same at any given point in time. Then, we show that for a given AU, the optimal camera setting changes depending on the environmental conditions i.e. different camera settings are optimal for an AU for different environmental conditions. After this, we systematically study the impact of optimal camera settings for different AUs and show optimal camera setting conflict i.e. optimal camera setting for one AU degrades performance of another AU. Later, we show that even with pre-recorded public video datasets, we observe similar behavior when multiple AUs are running simultaneously. 
%Furthermore, we show that post-capture image processing helps, but if the capture is of poor quality then post-processing won't be able to improve AU accuracy further and in such cases direct tuning of camera parameters is needed to improve AU accuracy. 
Finally, we show that there exists some camera setting that works well
for multiple AUs at the same time, which motivates us to study
automatically finding such a setting while the system
is in operation.

Since any video camera can have only one camera setting at any given
point it time, it is impossible to design experiments to study the effect
of different camera settings on different AUs at the same time, for
the same exact input scene. The closest alternative is to use multiple
cameras, with different setting for individual AUs. However,
there are thousands of possible camera settings, and
it is not practical to have as many cameras to cover all settings for
the same input scene.
As a work around, we chose two alternatives: (1)
Fix a static input scene in a controlled setting, point the video
camera to this scene and run multiple AUs with different camera
setting on this single camera. This way, we can obtain accuracy for
the AUs with different camera setting for the same input scene (in a
controlled environment).
(2) Use two side-by-side cameras, one with the
default camera setting and another with ``modified" camera setting,
and then recreate and repeat the exact same sequence of steps in the
scene, thereby providing similar input video content/scene for
different ``modified" camera settings, and measure the resulting AU
accuracy. In all our experiments, we modify four popular camera
settings, i.e., {\em brightness, contrast, color-saturation} (also known
as colorfulness), and {\em sharpness}, which are commonly found in
almost all video cameras. All other camera settings are kept at
manufacturer-provided default values.

%Analyzing the impact of camera settings on video analytics poses a significant challenge: it requires applying different camera settings to the same input scene and measuring the resulting accuracy of insights from an AU. The straightforward approach is to use multiple cameras with different camera settings to capture the same input scene. However, this approach is not practical as there are thousands of different possible camera settings and it is impossible to set up as many cameras to observe identical input scene. To overcome this challenge, we proceed with a workaround in which we place two side-by-side cameras, one with default camera setting and the other with ``modified" camera setting, and then recreate and repeat the exact input scene for different ``modified" camera settings, and measure the resulting AU accuracy. In our experiments, we considered four popular camera settings i.e. {\em brightness, contrast, color-saturation} (also known as colorfulness), and {\em sharpness}, which are commonly found in almost all video cameras. 

%\input{optimal_setting_within_AU}

%\subsection{Impact of Camera Settings on Different AU Accuracy}
\subsection{Optimal camera setting varies across AUs}
\label{subsec:optimal_setting_across_au}

\subsubsection{Different AUs have optimal camera settings}
\label{subsec:diff_optimal_setting}

\begin{figure}[tp]
    \centering
        \includegraphics[width=0.8\linewidth]{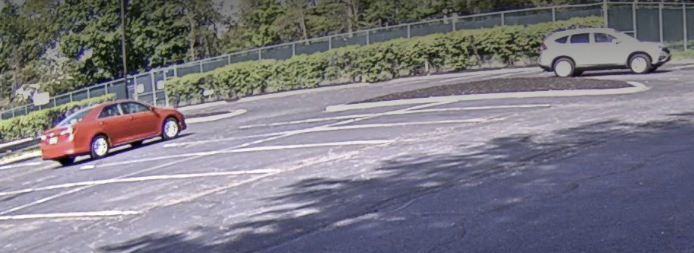}
        %\vspace{-0.1in}
        \caption{Scene for multiple AU experiments}
        \label{fig:experimental_scene}
\end{figure}

\begin{table}[tb]
\begin{center}
%\vspace{-0.15in}
\caption{Optimal camera settings for different AUs.}
%\vspace{-0.1in}
\label{tab:diff-aus}
{
\begin{tabular}{|c|c|c|}
    \hline
    AU & Optimal Camera Setting & Additional detections \\
    & [brightness, contrast] &  (over default setting)\\
    \hline
    FD & [30,80] & 89 \\
    \hline
   PD & [40,80] & 52 \\
    \hline
     CD & [65,30] & 2 \\
     \hline
     LPD & [80,65] & 14\\
    \hline
\end{tabular}
}
\end{center}
%\vspace{-0.1in}
\end{table}

In ~\cite{paul2021camtuner}, it is shown that optimal camera settings
for a \textit{single} AU is not the same all the time, i.e., it changes
as the environmental condition and video content change over time. In this
section, we show that the optimal camera settings is not the same for
\textit{different} AUs at any given point in time. Since we do not
have any public datasets that have identical videos captured with
different camera settings, we created our own video dataset. 
To collect a video in the dataset, we place two side-by-side cameras
at a parking lot and have two participants repeat the following
sequence of steps: drive the car towards the camera, show their face,
drive past the camera, park the car, walk towards the camera and walk
past the camera. The two side-by-side cameras used were Axis Q3505 MK
II Network camera, each capturing frames at 10 FPS and the scene
observed by the cameras is shown
in~\figref{fig:experimental_scene}. The sequence of steps followed by
the two participants allows us to perform multiple analytics, namely,
car detection (CD), person detection (PD), face detection (FD) and
license plate detection (LPD) on a single video stream and obtain
results for different ``modified" camera settings. In total, we
repeated the sequence of steps for 26 different camera settings and
observed the resulting AUs accuracy on each of the ``modified" camera
setting.

~\tabref{tab:diff-aus} shows that for the four AUs that run
simultaneously on the video with the same ``modified" camera setting,
each AU has a different camera setting that gives the highest accuracy
for the AU. For example, FD has highest accuracy with brightness set
to 30 and contrast set to 80, while CD has highest accuracy with
brightness set to 65 and contrast set to 30. 
All our results are normalized to the detections obtained with the
default camera setting, and the ``modified" camera setting which shows
the highest number of additional detections~\footnote{We manually
  verified that these additional detections were all true positive
  detections} over the default setting (shown in the right most column
in ~\tabref{tab:diff-aus}) is considered the optimal camera setting for
the specific AU. From ~\tabref{tab:diff-aus}, we can clearly see that
all four AUs have different optimal camera setting, i.e., the camera
setting that gives the highest accuracy for the AU. \textit{Thus,
  optimal camera setting varies across AUs}.

\subsubsection{Optimal camera settings conflict}
\label{subsec:optimal_settings_conflict}

In ~\secref{subsec:diff_optimal_setting}, we saw that different AUs
have different optimal camera setting at any given point in
time. Assume that there are two AUs, say AU$_1$ and AU$_2$ with
optimal camera settings as S$_1$ and S$_2$, respectively.
%Now, say AU$_1$ achieves accuracy of Acc$_1$ with S$_1$, while AU$_2$ achieves accuracy of Acc$_2$ with S$_2$. 
Now, since the camera can have only one setting at any given point in
time, what happens to A$_2$'s accuracy if the camera
setting is S$_1$ and vice versa, i.e., what happens to A$_1$'s accuracy if the
camera setting is S$_2$?
%
% Does the accuracy stay the same, improve or degrade ? 
To perform such a study, we set up two controlled experiments and show
below that there exists a conflict between the AUs, i.e., the optimal
camera setting for one AU degrades performance of another AU.

%In this section, we show that there is a conflict between these optimal camera settings for different AUs i.e. optimal camera setting for one AU degrades performance of another AU.

%and in ~\cite{paul2021camtuner} it is shown that even for the same AU the optimal camera setting varies depending on the environmental conditions. 
%In this section, we show the conflict in optimal camera settings when multiple AUs are run simultaneously. 

For the first controlled experiment, we print 12 different person
cutouts obtained from the COCO dataset~\cite{lin2014microsoft} and place
them in front of a real camera. Next, we use two sources of light. One
is always turned ON and the other is turned ON to simulate DAY
condition and the turned OFF to simulate NIGHT condition.
%The observed scene by the camera in default camera settings is shown in ~\figref{fig:day-night-capture} for DAY and NIGHT conditions. 
By using two sources of light, we can change the environmental
conditions around the camera by just turning one source of light ON or OFF. 
In addition, we set up a controlled experiment with 3D models of
people and cars being observed by the AXIS Q3515 camera and then show the
conflict between PD, FD and car detection (CD) on this second setup.

%Later, we show that such conflict is observed even with pre-recorded public video datasets, when multiple AUs are run simultaneously. 

%\subsubsection{Real cameras}
%\label{subsec:optimal_setting_conflict_real_cameras}

\begin{table}[tb]
\centering
\caption{Optimal camera settings conflict (2D printouts).}
 \begin{subtable}[tb]{0.45\textwidth}
\centering
\begin{tabular}{|c|c|c|c|}
    \hline
    AU &  Optimal camera setting &  FD accuracy & PD accuracy \\
    & [brightness, contrast & mAP (\%) & mAP (\%)\\
    &  color, sharpness] & & \\
    \hline
    PD & [40,90,60,100] & 82.58 & \textbf{99.36} \\
    \hline
    FD & [60,40,90,90] & \textbf{100.0} & 83.33 \\
    \hline
\end{tabular}
\caption{DAY}
\label{tab:day-best-cam}
\end{subtable}
\begin{subtable}[tb]{0.45\textwidth}
\centering
\begin{tabular}{|c|c|c|c|}
    \hline
    AU &  Optimal camera setting &  FD accuracy & PD accuracy \\
    &  [brightness, contrast & mAP (\%) & mAP (\%)\\
    &  color, sharpness] & & \\
    \hline
    PD & [80,90,70,100] & 38.89 & \textbf{83.33} \\
    \hline
    FD & [80,90,60,80] & \textbf{50.0} & 75.0 \\
    \hline
\end{tabular}
\caption{NIGHT}
\label{tab:night-best-cam}
\end{subtable}
\label{tab:day-night-best-cam}
 \vspace{-0.1in}
\end{table}

%\begin{table}[tb]
%\centering
% \begin{subtable}[tb]{0.45\textwidth}
%\centering
%\begin{tabular}{|c|c|c|c|}
%    \hline
%    AU &  Optimal camera setting &  FD accuracy & PD accuracy \\
%    &  & mAP(\%)/IoU & mAP(\%)/IoU\\
%    \hline
%    PD & [40,90,60,100] & 82.58/0.66 & 99.36/0.81 \\
%    \hline
%    FD & [60,40,90,90] & 100.0/0.66 & 83.33/0.87 \\
%    \hline
%\end{tabular}
%\caption{DAY}
%\label{tab:day-best-cam}
%\end{subtable}
%\begin{subtable}[tb]{0.45\textwidth}
%\centering
%\begin{tabular}{|c|c|c|c|}
%    \hline
%    AU &  Optimal camera setting &  FD accuracy & PD accuracy \\
%    &  & mAP(\%)/IoU & mAP(\%)/IoU\\
%    \hline
%    PD & [80,90,70,100] & 38.89/0.66 & 83.33/0.78 \\
%    \hline
%    FD & [80,90,60,80] & 50.0/0.74 & 75.0/0.78 \\
%    \hline
%\end{tabular}
%\caption{NIGHT}
%\label{tab:night-best-cam}
%\end{subtable}
%\caption{Optimal camera settings conflict}
%\label{tab:day-night-best-cam}
% \vspace{-0.3in}
%\end{table}

~\tabref{tab:day-night-best-cam} shows the conflict when we apply
optimal camera settings of PD to FD and vice versa for DAY and NIGHT
conditions. For DAY condition, 
% as seen in ~\tabref{tab:day-best-cam},
the optimal camera setting for PD (brightness: 40, contrast: 90,
color: 60, sharpness: 100) gives 99.36 \% accuracy for PD but the same
camera setting gives only 82.58 \% accuracy for FD, which with another
camera setting, i.e., brightness: 60, contrast: 40, color: 90,
sharpness: 90, achieves 100 \% accuracy. 
Conversely, when we
apply optimal camera setting for FD to PD, then the accuracy of PD
drops from 99.36 \% to 83.33 \%. Similar observation holds for NIGHT
condition as well, as shown in ~\tabref{tab:night-best-cam}. Here,
the optimal setting for FD achieves 50 \% accuracy, but the accuracy drops to
38.89 \% accuracy if the optimal setting for PD is applied to FD. Also,
the optimal setting for PD achieves 83.33 \% accuracy but the accuracy drops to 75 \% if
the optimal setting of FD is applied to PD.

\begin{figure}
    \centering
        \includegraphics[width=0.8\linewidth]{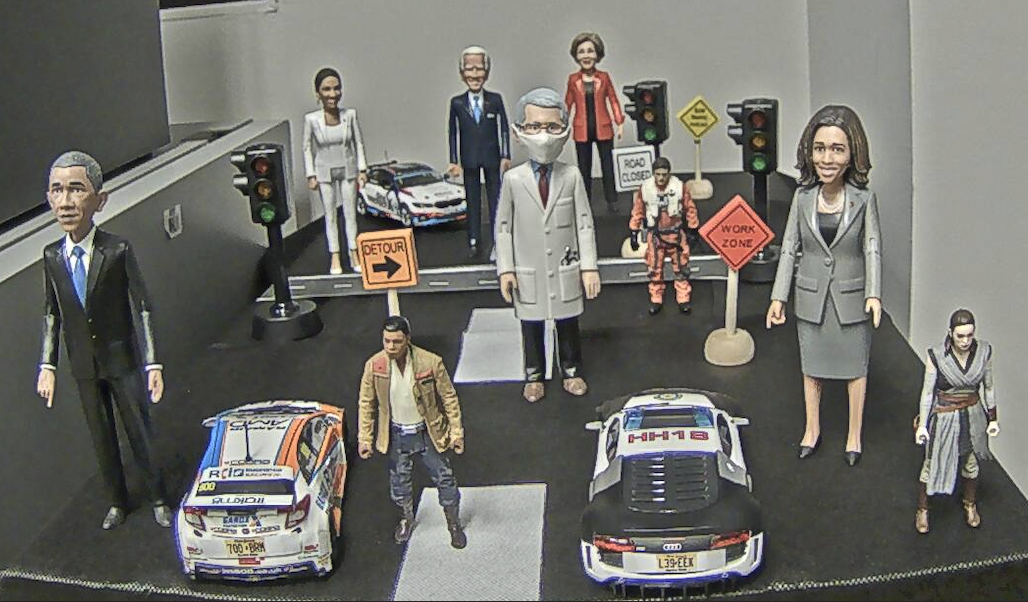}
        \caption{Controlled experiment with 3D models.}
        \label{fig:demo_scene}
%\vspace{-0.1in}
\end{figure}

\begin{table}[tb]
\begin{center}
\vspace{-0.15in}
\caption{Optimal camera settings conflict (3D models).}
    %\vspace{-0.1in}
\label{tab:demo3d}
{
\small
\begin{tabular}{|c|c|c|c|}
    \hline
    Optimal camera setting & FD & PD & CD  \\\relax
    [brightness,contrast, & accuracy & accuracy & accuracy \\
    color,sharpness] & mAP (\%) & mAP (\%) & mAP (\%)\\
    \hline
    FD & \textbf{52.1} & 78.7 & 54.5\\\relax
     [60,30,40,80] & & & \\
    \hline
   PD & 37.3 & \textbf{100.0} & 66.7 \\\relax
    [80,60,50,40] & & & \\
    \hline
     CD &  41.9 & 85.8 & \textbf{96.9} \\\relax
     [70,50,60,60] & & & \\
    \hline
\end{tabular}
}
\end{center}
\vspace{-0.1in}
\end{table}

%In addition to 2D printed person cutouts, we also setup another controlled experiment with 3D models of people and cars, as shown in ~\figref{fig:demo_scene}. 

For the second controlled experiment, we set up 3D models of people and
cars as shown in ~\figref{fig:demo_scene}. Here, we modify the values of four
camera parameters, i.e., brightness, contrast, color and sharpness, 
each from 0 to 100 in steps of 10, and capture the image from the camera
for each of the ``modified" setting. In total, we obtain
$\approx$14.6K (11$^4$) different images corresponding to different
camera settings and run three different AUs, i.e., PD, FD and CD on each
of these images.
~\tabref{tab:demo3d} shows the conflict observed for
these three AUs. The optimal setting for FD achieves 52.1 \% accuracy, but
the accuracy drops to 37.3 \% when the optimal setting of PD is applied to FD, and
to 41.9 \% when the optimal setting for CD is applied to
FD.
Similarly, PD achieves 100 \% accuracy under its own optimal setting, 
but only 78.7 \% accuracy when the optimal setting for FD is applied , and
85.8 \% when the optimal setting of CD is applied.
As with FD and PD, conflict is also observed for CD.  The CD accuracy
is 96.9 \% under its own optimal setting, but drops to 66.7 \% and
54.5 \% when the optimal settings for PD and FD are applied,
respectively. \textit{Thus, the optimal camera settings of AUs
conflict with each other when multiple AUs are run simultaneously.}

\subsection{Common-optimal camera settings exists}
\label{subsec:optimal_setting_exists}

\begin{table}[tb]
\caption{Common good camera settings across AUs.}
\begin{subtable}[tb]{0.75\linewidth}
\centering
\resizebox{\textwidth}{!}{%
\begin{tabular}{|c|c|c|c|c|}
    \hline
     Camera Setting & FD accuracy & PD accuracy\\
    & mAP (\%) & mAP (\%)  \\
    \hline
    Default & 82.58 & 90.9  \\
    \hline
    PD-optimal & 82.58 & 99.36  \\
    \hline
    FD-optimal & 100.0 & 83.33  \\
    \hline
    \emph{Common-optimal} & 100.0 & 91.7 \\
    \hline
\end{tabular}}
\caption{DAY}
\label{tab:day-overall}
\end{subtable}

\begin{subtable}[tb]{0.75\linewidth}
\centering
\resizebox{\textwidth}{!}{%
\begin{tabular}{|c|c|c|c|c|}
    \hline
    Camera Setting & FD accuracy & PD accuracy \\
    & mAP (\%) & mAP (\%) \\
    \hline
    \emph{Default} & 38.89 &  58.3 \\
    \hline
    PD-optimal & 38.89 & 83.38 \\
    \hline
    FD-optimal & 50.0 & 75.0 \\
    \hline
    \emph{Common-optimal} & 48.8 & 83.3 \\
    \hline
\end{tabular}}
\caption{NIGHT}
\label{tab:night-overall}
\end{subtable}
\label{tab:overall-optimal}
 \vspace{-0.2in}
\end{table}

In ~\secref{subsec:diff_optimal_setting} we saw that different
AUs have different optimal camera setting at any given point in time,
%in ~\secref{subsec:optimal_setting_within_au} we saw that even for the same AU the optimal camera setting varies depending on the environmental conditions 
and in ~\secref{subsec:optimal_settings_conflict} we saw that optimal
camera setting across AUs conflict with each other, i.e., the optimal
setting for one AU degrades performance of another AU. In this
section, we show that there exists a common-optimal camera setting that works well across multiple
AUs. This motivates us to automatically find 
%such 
this common-optimal
camera setting during online operation of the video analytics system.

We use the same setup as discussed
in~\secref{subsec:optimal_settings_conflict} with two sources of
light. ~\tabref{tab:overall-optimal} shows the accuracy for FD and
PD for different camera settings, i.e., the default camera setting, optimal
camera setting for PD (PD-optimal), optimal camera setting for FD
(FD-optimal), and common-optimal camera setting
(\textit{Common-optimal}) for DAY and NIGHT conditions.
%As seen earlier in ~\tabref{tab:day-night-best-cam}, we see conflict when optimal setting of one AU is applied on another AU. 
In this experiment, we define the common-optimal setting
as the one  that gives the highest cumulative 
accuracy of the two AUs (PD and FD). Section~\ref{subsec:rl}
discusses alternative definitions of common-optimal setting
according to how the accuracy estimates of multiple AUs are aggregated.

%  and choose that as the common
% optimal camera setting for DAY and NIGHT condition. 
% (among the $\approx$14.6K images, we picked the camera settings 

~\tabref{tab:day-overall} shows that with common-optimal camera
setting, we are able to restore the drop in FD accuracy from 100 \% to
82.58 \%, back to 100 \%. Also, due to the conflict, PD accuracy
dropped from 99.36 \% to 83.33 \%, but with the common-optimal camera
setting, we are able to bump it up to 91.7 \%. We see similar
improvement in accuracy with common-optimal camera settings for NIGHT
condition as well, as shown
in~\tabref{tab:night-overall}. \textit{Thus, there exists a
  common-optimal camera setting, which
%resolves the conflict to some extent and this camera setting 
works well across multiple AUs}.
%\begin{figure}
%    \centering

        % \includegraphics[width=\textwidth]{figs/impact_all_2.pdf}
%        \includegraphics[width=0.5\textwidth]{figs/multiAU_setup.PNG}
%        \caption{Single Camera Setting and Multiple Virtual Camera Settings for Multiple AU setup}
%        \label{fig:mulAU_setup}
%\vspace{-0.1in}
%\end{figure}

%\textcolor{red}{Although camera parameter tuning provides superior performance over post-capture image transformation. In a multi-AU secenario, the overall performance can further be boosted by applying each AU-specific image transformation on the frame captured with overall best camera setting based on the combined perception of multiple AUs.} Although each AU has its own best camera setting and these best camera settings are suboptimal for other AUs but there exists a camera setting that provides overall best performance for multiple AUs. ~\tabref{tab:overall-optimal} shows the overall best cumulative AU accuracy (i.e., sum of individual AU accuracy) for overall-optimal camera setting. 

%~\tabref{tab:overall-optimal} also indicates that applying different digital transformation adapting to various AUs' perception can also enhance the AU performance. ~\figref{fig:mulAU_setup} shows how multiple AU can have its own virtual knob settings while camera is tuned at a overall-optimal camera setting.

\section{Challenges and Approaches}
\label{section:challenge}

In this section, we describe the key challenges that we face in order
to alleviate the ill-effects of one AU camera setting on another AU,
and improve the accuracy of multiple AUs (running on a single camera
stream) simultaneously.
%enable multiple AUs to run on a single video stream with high accuracy.

%\subsection{Independent challenges}

%\subsection{Approach-dependent challenges}

{\bf Challenge 1: How to define the common-optimal camera setting?} 
As discussed in Section~\ref{sec:motivation}, the camera setting has a
direct impact on the accuracy of AUs and when there are multiple AUs
running on a single video stream, depending on the camera setting,
some AUs may perform better while others perform worse.
%then this behavior might change and previously under-performing AU may now start performing better than the previously better-performing AU.
To improve accuracy of multiple AUs simultaneously on a single video
stream, we need to choose a camera setting that optimizes the
aggregate accuracy for all AUs, a.k.a. the common-optimal camera
setting. Depending on the AU, the acceptable accuracy may vary and for
certain AUs, accuracy improvement over a certain threshold may not be
very significant. Therefore, it is not so straight forward to define
the common-optimal camera setting, which gives optimal performance
across multiple AUs.

%Now, in order to improve performance of all AUs simultaneously, we need to choose the camera setting that is ``good" for all the AUs. We refer to this camera setting as the common-optimal camera setting. Defining a ``good" camera setting for all AUs is tricky, because there are multiple ways in which this can be defined e.g. a ``good" camera setting can be the one that gives highest average accuracy across AUs or a ``good" camera setting can be the one that ensures all AUs perform at a certain minimum accuracy. Depending on the AU, the acceptable accuracy may vary and for certain AUs, accuracy improvement over a certain threshold may not be very significant. Therefore, it is not so straight forward to define the common-optimal camera setting.

%\textbf{Approach.} we define the template of the objective function where users can set the coefficients of priorities. 

\textbf{Approach.} Since the meaning of the change in accuracy numbers
varies across AUs and there is no generic method to determine
``optimal" accuracy for any given AU, we choose to define an objective
function that aggregates the accuracy for multiple AUs into a single
metric. The objective function allows for different aggregation strategies,
which are defined by domain experts, and we design \approach\ to choose
the camera setting that optimizes the aggregate objective function,
which works the best across multiple AUs.

%and provide a template for determining the common-optimal camera setting, where the co-efficients in the objective function is provided by a domain expert, who understands the meaning of the relative changes in the accuracy numbers for the various AUs.

{\bf Challenge 2: How to achieve common-optimal camera settings for a
  particular scene for multiple AUs?} Defining the common-optimal camera
setting is one thing, but achieving the common-optimal camera setting
is whole another thing. As the scene in front of the camera changes
(due to environmental changes, lighting condition changes or change in
flow of people/objects, etc.), the accuracy of AUs also changes, and thus
the common-optimal camera setting for the particular
scene also changes. This change of scene is inevitable and continuous, 
and it is quite challenging
to adapt to such changes and achieve the common-optimal camera setting
for a particular scene for multiple AUs in real-time.

%Therefore, depending on the scene the common-optimal camera setting changes and achieving this change in real-time as the scene changes is quite challenging.

%\textbf{Approach.} Since we can have only one physical setting at camera at any given time. use of online learning is only feasible option

\textbf{Approach.} One way to find the common-optimal camera setting
for a particular scene for multiple AUs is to train a model that knows
the various scenes and for the specific combination of the AUs, so
that it can provide the common-optimal camera setting for any given
scene. However, training such a model is almost impractical, since we
do not know a priori the various conditions that would be observed by
the cameras at various locations. Therefore, our approach is to use
online learning, where the common-optimal camera setting for the
specific AUs is learned in an online fashion, after the camera is
deployed.

{\bf Challenge 3: No Ground truth during online operations.} Implicit to the
definition of the common-optimal camera setting is the accuracy of
individual AUs. To determine the accuracy of AUs, it is essential
to know the ground truth, so that one can compare the results with the
ground truth and determine the accuracy. However, knowing such ground
truth during online operations is not possible, since there is no human in the loop to
annotate the frames in real-time. In such a scenario, how can we
determine the common-optimal camera setting for multiple AUs running
on a single video stream?

%\textbf{Approach.} proxy for quality evaluation of analytics outcome in absence of ground-truth

\textbf{Approach.} Since we do not have actual ground truth in
real-time and we cannot determine the exact accuracy of AUs, we choose
to \textit{estimate} the accuracy of the AUs using proxy, similarly as
in~\cite{paul2021camtuner}. Since each AU has a different method of
measuring accuracy, our proxy is also different for each AU
and we keep it light-weight, so that we can \textit{estimate} the
accuracy in real time. By using light-weight, AU-specific
quality estimator, we compensate for the lack of availability of
ground truth in real-time.

%{\bf Challenge 4: Common best is not the best for each AU. Thus, how to improve AU accuracy over common best.}
%\textcolor{blue}{Maybe take out the below challenge 4 ?}

%{\bf Challenge 4: common-optimal camera setting does not give best accuracy for each AU}. common-optimal camera setting, by definition, caters to multiple AUs and is not able to achieve the best possible accuracy for each individual AU. Given that we know it is possible to improve the AU accuracy further i.e. after setting the camera to the common-optimal setting, what can we do to enhance the accuracy further and achieve the best accuracy for each AU, when they are running simultaneously on a single video stream ?

%\textbf{Approach.} AU-specific post-processing on top of common-optimal camera setting. 2 avenues for improving AU accuracy. if camera setting is becoming better scope of improvement will lower down

%\textbf{Approach.} We realize that we can improve the AU accuracy by improving the quality of data that we feed to the AU. Now, data quality can be improved at the source i.e. by changing camera parameters and it also can be improved with post processing. Thus, our approach to achieve best possible accuracy for individual AU is as follows: after changing the camera setting to common-optimal setting, on top of that, we perform AU-specific post processing on the input data i.e. frame, such that the quality of input data further improves, which results in improvement in individual AU accuracy.

%{\bf Challenge 5: Extremely slow RL training}
%\textcolor{blue}{Shall we keep or take out below challenge 5 ?}

{\bf Challenge 4: Extremely slow online learning}. Online learning
helps us to achieve common-optimal camera settings for a particular scene
for multiple AUs. However, the learning is extremely time
consuming. This is because, it takes a very long time to undergo
various conditions and as part of online learning, every change in
camera settings takes a long time ($\sim$ 200 ms). This limits the
speed of online learning to the point that it can become ineffective.

\textbf{Approach.} We pre-learn (offline) the values of various
AU-specific estimates for different AUs depending on the measured
values of brightness, contrast, color and sharpness for the images and
corresponding camera setting values. This helps in accelerating the
learning process when the camera is deployed in operation, because the
online learning agent is already bootstrapped with some pre-computed
values which allows it to start exploration and exploitation immediately.

\vspace{-0.1in}
\section{Design}
\label{sec:design}

\begin{figure*}[t]
   \centering %\includegraphics[width=0.96\linewidth]{figs/elixir_design_1.PNG}
   \includegraphics[width=0.98\linewidth]{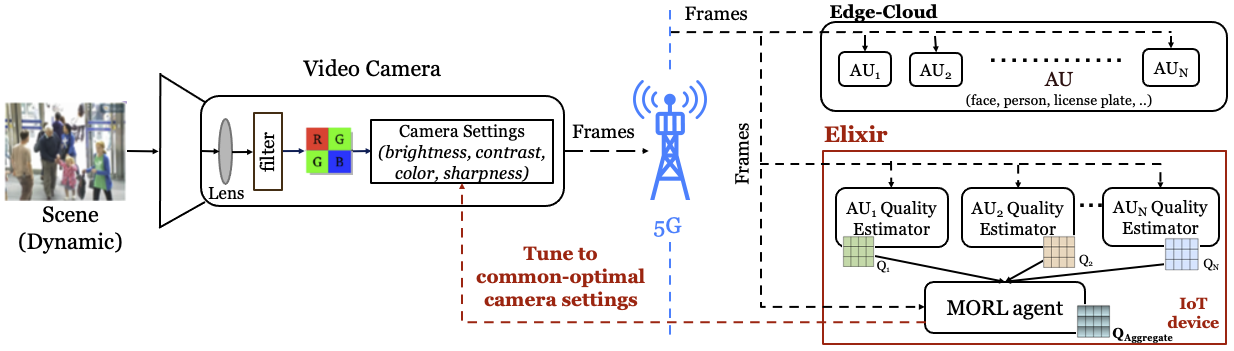}
   \caption{$\approach$ system design.} 
   %\vspace{-0.1in}
\label{fig:system_design}
   \vspace{-0.1in}
\end{figure*}

In this section, we discuss \approach's overall system design and its
two key components, the Multi-Objective Reinforcement Learning (MORL)
agent and AU-specific Quality Estimators, which aid in dynamically
finding the common-optimal camera settings that enhances the input
video quality, such that the accuracies of multiple AUs running on a
single video stream are improved simultaneously.

~\figref{fig:system_design} shows the overall system design for
\approach. As the scene in the field of view of the camera changes
dynamically, the video camera captures the input scene and depending
on the camera settings, emits frames, which are then transmitted over
5G network to different AUs for further processing. In parallel,
frames are also transmitted to the \textit{AU-specific quality
  estimators} and the \textit{MORL agent}. AU-specific quality
estimators provide estimates for the quality of the frames that correlates with 
the AU-specific accuracy; the higher the quality, the better chances
the AU will give high accuracy on the frame. 
The MORL agent obtains the quality estimates from individual
AU-specific quality estimators, combines them into an aggregate
single value for the frame, and learns to tune the camera settings to a
common-optimal setting that works well for multiple AUs
simultaneously. 
Next, we discuss the two key components, MORL
and AU-specific quality estimators, in detail.

%, which then analyze the frames and tune the camera settings to common-optimal settings so that accuracy of all AUs is improved simultaneously. 

%~\figref{fig:system_design} shows the system-level architecture for \approach\ that continuously and dynamically tunes four camera parameters (\ie Brightness, contrast, color, sharpness) to enhance the accuracy of multiple downstream AUs in the VAP. \approach has two key components: a Multi-Objective Reinforcement Learning (MORL) engine, and AU-specific analytics quality estimator.

%\subsection{Multi Objective Reinforcement learning (MORL) Engine}
\subsection{Multi-Objective Reinforcement Learning}
\label{subsec:rl}

\begin{figure}[tb]
    \centering
        \includegraphics[width=0.99\linewidth]{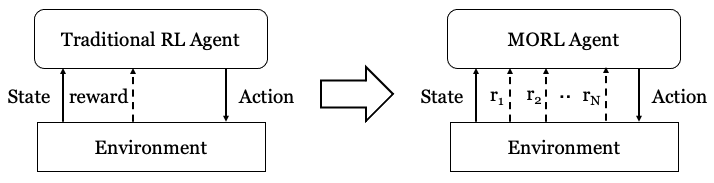}
        \caption{Traditional RL vs MORL architecture}
        \label{fig:traditional_vs_morl}
        \vspace{-0.2in}
\end{figure}

Reinforcement Learning (RL) is a popular technique for online learning
where the agent observes the environment (\textit{state}) that it is
in, performs an \textit{action} among the allowed set of actions and
learns to maximize a \textit{reward} function, which is set up in
order to achieve the desired goal. In ~\cite{paul2021camtuner}, only a
\textit{single} AU was considered, and therefore, a single-objective,
traditional RL technique could be used to tune the camera settings to
improve accuracy of a single AU. However, for \approach, there are
\textit{multiple} AUs running simultaneously and the goal is to
improve the accuracy for all of them simultaneously. Hence, we use a
Multi-objective Reinforcement Learning
(MORL)~\cite{liu2014multiobjective,morl-2} technique with multiple
objectives (one for each AU) to find the
common-optimal camera setting that enhances the data quality such that
the accuracies of multiple AUs are improved simultaneously.

~\figref{fig:traditional_vs_morl} shows the difference between
traditional RL and MORL. In traditional RL, the agent receives only a
single reward, whereas in MORL the agent receives multiple rewards,
one corresponding to each objective. The goal of the MORL agent is to
improve all these individual rewards. In our case, we have multiple
AUs running off of a single video stream and the goal is to improve
accuracy for each of these AUs. This accuracy for each AU is
determined by the reward that the agent obtains for each AU. Now, if
the action taken by the agent improves the accuracy of the AU, then
the reward is positive, otherwise it is negative. \approach\ uses
AU-specific quality estimators (discussed
in~\secref{subsec:au-quality}) as reward function for each AU.

Next, we define the \textit{State}, \textit{Action} and the
\textit{Reward} used by the MORL agent in \approach.

\textit{\underline{State}}: State is a tuple of two vectors,
$s=<P_{t}, M_{t}>$, where $P_t$ consists of the values of brightness,
contrast, color and sharpness in the camera settings at time $t$, and
$M_{t}$ consists of the measured values of brightness, contrast, color
and sharpness of the captured frame at time $t$.

\textit{\underline{Action}}: The set of actions for the MORL agent are
(a) to increase or to decrease camera setting value for brightness,
contrast, color or sharpness (only one among the four) (b) no change
to any camera setting values. Thus, in total, there are 9 actions --
increasing or decreasing one of the four camera settings 
(8 actions) and no change (the 9\textsuperscript{th} action).

\textit{\underline{Reward}}: Value of an aggregate function that
combines the values of AU-specific quality estimators for all the AUs
running off of the single video camera stream.

\begin{algorithm}[tb]
%\DontPrintSemicolon
\SetKwFunction{OE}{Observe-Environment}
\SetKwFunction{CA}{Choose-Action}
\SetKwFunction{Agg}{Aggregate}
\SetKwFunction{SP}{Still-Processing}
\SetKwFunction{PA}{Perform-Action}
\SetKwFunction{CR}{Compute-Reward}
\SetKwData{S}{$s$}
\SetKwData{SPRIME}{s'}
\SetKwData{A}{a}
\SetKwData{APRIME}{a'}
\SetKwData{Q}{Q}
\SetKwData{TQ}{Q$_{aggregate}$}
\SetKwData{Qi}{Q$_i$}
\SetKwData{ri}{r$_i$}
\SetKwData{AUi}{AU$_i$}
\SetKwData{N}{N}
\SetKwData{K}{K}
\SetKwData{R}{r}
\SetKwData{ALPHA}{alpha}
\SetKwData{GAMMA}{gamma}
\SetKwInOut{Input}{Input}

\Input{\N = Number of AUs/objectives}
\Input{\K = Number of episodes}

Initialize State \S\\
Initialize \Q table for each AU and \TQ\ (N+1 Q-tables)\\
\tcp{During exploration upto episode K}
\While{\SP{}}{
\A $\leftarrow$ \CA{\TQ, \S}\\
\PA{\A}\\
\SPRIME $\leftarrow$ \OE{}\\
\For{AU i = 1, 2, ..., \N}{
\tcp{AU-specific quality estimation}
\ri $\leftarrow$ \CR{}\\
\tcp{Bellman optimality equation}
\Qi(\S,\A) $\leftarrow$ \Qi(\S,\A) + $\alpha$ $\times$ [\ri + $\gamma$ $\times$ \Qi(\SPRIME, \APRIME) – \Qi(\S, \A)] \\
}
\tcp{optimal aggregation policy}
\TQ(\S,\A)  $\leftarrow$ \Agg(\Q(\S,\A))\\
\S $\leftarrow$ \SPRIME\\
}
\caption{Single-policy approach to MORL}
\label{alg:MORL}
\end{algorithm}

Several methods can be used by the MORL agent in 
using the individual reward functions and ``learning" to take appropriate
actions. The learning process by the agent leads to a ``policy", which
guides the agent in taking actions leading to common-optimal camera
settings.
Now, there can be ``single-policy" or ``multiple-policies" algorithms
that can be used in MORL~\cite{liu2014multiobjective,morl-2}. As
pointed out in~\cite{morl-2}, ``single-policy'' algorithms are more
appropriate when the aggregation function, a.k.a. utility function or
scalarization function, is known apriori and does not change over
time, whereas ``multi-policy" algorithms are more appropriate when the
aggregation function is not known or changes over time. 
In \approach, we define our own aggregation function which is not
changed over time. Thus, ``single-policy" algorithm is preferable by
the MORL agent in \approach. In this method, the MORL agent addresses
the multi-objective problem by using the aggregation function to reduce
the multi-objective problem into a single-objective problem, where
the individual reward values are aggregated into a single reward
value. The MORL agent then goes after this ``aggregated" single reward
value and ``learns" to maximize it over time. 

Algorithm~\ref{alg:MORL}
shows the steps followed by the MORL agent in ``single-policy"
learning algorithm. Here, we first initialize the state $s$ and (N+1)
Q tables: N Q-tables corresponding to N AUs and final one is the
aggregated Q table, Q$_{aggregate}$.
%We always choose the action to perform on the camera from Q$_{aggregate}$ table based on the current state following a $\epsilon$-greedy policy. 
After initialization, the various iterations, a.k.a. episodes, start for
the MORL agent. The first step is to choose an action to perform.  We
follow an $\epsilon$-greedy policy to choose an action based on the
the current state and values in the Q$_{aggregate}$ table. 
The next step is to
perform the action, and after performing the action on camera (\ie
changing the camera settings), the MORL agent observes the next modified
state, s', and computes the immediate reward for each AU (discussed
later in ~\secref{subsec:au-quality}). 
Following this, for each AU, the MORL agent updates the corresponding
Q-table entry independently using the \emph{bellman optimality
  equation}. In this equation, $\alpha$ is the learning rate (between
0 and 1) and $\gamma$ is the discount factor (between 0 and
1). $\alpha$ controls the importance given to new information. 
The higher the value, the more importance is given to most recent
information. $\gamma$ controls the priority given to long-term reward
or immediate reward. When $\gamma$ is 1, the MORL agent  values
long-term rewards highly, whereas when the value is 0, the MORL agent
completely ignores long-term rewards and optimizes for immediate
rewards. 
Once the Q-table entry for each AU is computed, the MORL
agent then uses an aggregation function to combine the Q-table values
of each AU into an aggregate value, which is updated in
the Q$_{aggregate}$ table. Finally, the state is updated to s' and the
particular iteration/episode ends.
%How much importance, MORL agent gives to new information, is controlled by $\alpha$. MORL agent prioritizes the long-term reward or immediate reward based on a constant, $\gamma$ (\ie also known as discount factor). We then use the aggregation function (\ie discussed below) to update the Q$_{aggregate}$ table.  

%%%%

The MORL agent performs exploration and exploitation alternatively.
In the $\epsilon$-greedy policy, the value of
$\epsilon$ (between 0 and 1) controls the balance between exploration
and exploitation. During exploration, we run algorithm~\ref{alg:MORL}
for fixed K episodes to populate the Q-tables for a few hours. The MORL
agent generates a random number between 0 and 1. If the random
number is greater than the set value of $\epsilon$, a random action
is chosen; otherwise the ``best" action for the state, i.e., the one
with the highest Q-value in the Q$_{aggregate}$ table is chosen. As we can
see, a lower value of $\epsilon$ will trigger more random actions to be
generated while a higher value of $\epsilon$ will trigger fewer random
actions. The MORL agent uses a smaller $\epsilon$ (\ie 0.1) during
exploration and a larger $\epsilon$ (\ie 0.9) during
exploitation. Also, during exploration, the learning rate,
$\alpha$ is set to be high (\ie 0.8), and during exploitation, it is set
low (\ie 0.2) in order to assimilate new information during
exploration and then use it during exploitation.

%Similar to $\epsilon$, learning rate, $\alpha$ is set to high (0.8) during exploration and low (0.2) during exploitation phase.

For the aggregation function used by the MORL agent, we considered
three different aggregation strategies as discussed below:

\begin{enumerate}
    \item \textbf{Strategy 1 (linear)}: This strategy gives equal
      weightage to all AUs. The aggregate function, as shown in
      Equation \ref{eq:linear}, is computed by taking the average of
      the reward functions (Q-values) obtained from the different AUs.
    \begin{equation}
        Q_{aggregate} = \sum_{i \in AU} Q_{i}[s,a] / count_{AU}
        \label{eq:linear}
    \end{equation}
        \item \textbf{Strategy 2 (weighted)}: 
          This strategy gives different weights to different AUs
          according to the priorities of the AUs and computes the
          aggregate value by using the below Equation 
          \ref{eq:weighted}.
        
        %It uses pre-defined non-uniform priorities (\ie weightages) based on granularity of the downstream task. We used the insight of face-detection and license plate detection is much more fine-grained than person and car detection. We assign priority values for four AUs \ie $p_{FD}$, $p_{LPD}$, $p_{PD}$, $p_{CD}$ as 3, 2, 1,1 respectively. 
    \begin{equation}
        Q_{aggregate} = \sum_{i \in AU} (p_{i} * Q_{i}[s,a])/\sum_i p_{i} 
        \label{eq:weighted}
    \end{equation}
        Such weights will be given by domain experts who understand the
        meaning and relative importance of the reward values for different AUs.
%  and the relative         importance that needs to be given to each one of them.
        %\comment{give the priorities used for PD, FD, LPD, etc. in the evaluation section.}
    
    \item \textbf{Strategy 3 (winner-takes-all)}: The maximum reward
      (obtained using the Bellman equation) among the different rewards
      from multiple AUs, is used in this aggregation strategy, i.e., the
      winner takes all, as shown in Equation
      \ref{eq:winner_takes_all}. Such a strategy ensures that the
      selected action is guaranteed to be optimal for at least one
      objective/AU.
    
    %Here the multiAU Q table is updated based on the maximum discounted reward (using bellman equation) corresponding to each downstream AU. Policy 3 ensures that the selected action is guaranteed to be optimal for at least one objective.
    %Under single-policy approach, here we use W-learning approach~\cite{humphrys1996action} where 
    \begin{equation}
        Q_{aggregate} = \max_{i \in AU} Q_{i}[s,a]
        \label{eq:winner_takes_all}
        \end{equation}
\end{enumerate}

\subsection{AU-specific quality estimators}
\label{subsec:au-quality}

The MORL agent in \approach\ receives individual reward functions
corresponding to the multiple AUs that are running off of a single
video stream, as discussed in~\secref{subsec:rl}. The reward function
for each AU signifies whether the action taken by the MORL agent is
improving or degrading the performance of the specific AU. A positive
reward indicates improvement, whereas a negative reward (penalty)
indicates degradation. During online operation, measuring the accuracy
of the AU is not possible because accuracy measurement requires
knowledge of the ground truth and such ground truth cannot be obtained
in real-time. As a proxy to this accuracy measurement, the concept of
AU-specific quality estimator was introduced
in~\cite{paul2021camtuner}, which \textit{estimates} the accuracy of
the AU. We draw inspiration from this technique and use a similar
approach in \approach, where we develop and use new AU-specific
quality estimators for the different AUs.

Specifically, the AU-specific quality estimator
in \approach\ consists of (1) feature extractor and (2) AU-specific
quality classifier. For the feature extractor, we use the
Inception-v3~\cite{inception-v3} network up to the first inception
module in order to accommodate the diverse impact of local
textures. These extracted features are used by the AU-specific quality
classifier to \textit{estimate} the accuracy of the AU. We use 2-fully
connected layers with 101 output classes (each output label
corresponds to mAP score between 0 to 100) for the classifier.

The four AUs that we use to demonstrate the effectiveness of Elixir 
work are: face detection (FD),
person detection (PD), car detection (CD) and license plate detection
(LPD). For each of these AUs, we need to develop a corresponding AU-specific
quality estimator.
%For FD, we follow the method used in ~\cite{paul2021camtuner} and obtain AU-specific quality estimator. 
Although both PD and CD fall in the category of ``object detection'',
unlike~\cite{paul2021camtuner}, we developed new AU-specific quality
estimator for each of the ``object'' classes, as there could be
inter-class variability. Specifically, we developed separate
AU-specific quality estimator for PD and CD. In addition, we developed
new AU-specific quality estimators for FD and LPD.

For the development of these new AU-specific quality estimators, we
used different public datasets. We picked 8000 frames from videos in
Roadway~\cite{canel2019scaling} and LSTN~\cite{fang2019locality}
datasets and created $\sim$ 4 million variants of them by applying
digital transformation on these frames~\footnote{We use the python-pil
  library~\cite{pil} for digital transformation}. These 4 million
frames constitute the training dataset for FD and PD. For CD, through
digital transformation on 6000 video frames extracted from Jackson and
Roadway datasets~\cite{canel2019scaling}, we created a training dataset
of $\sim$ 3 million frames. For LPD, we used 700 images from
Cars~\cite{krause20133d} and AOLP datasets~\cite{hsu2012application}
and created 0.6 million variants using digital transformation 
as the training dataset.

As in~\cite{paul2021camtuner}, we use a cross-entropy loss
function to train the AU-specific quality estimators, initial learning
rate is $10^{-5}$, and we use Adam Optimizer~\cite{kingma2014adam}.

\vspace{-0.1in}
\section{Evaluation}
\label{section:eval}

%\subsection{End-to-end VAP Performance}
%\label{subsec:mainresult}

We present our experimental results in this section. First, we present
the results of the various AU-specific quality estimators that we
developed. Then, we present the results for the different aggregation
strategies considered in MORL. Next, we present the results for real-world
deployment with three cameras placed next to each other 
running \approach\ and two alternative approaches. Finally, we
present some system performance results for \approach.

%We first evaluate the effectiveness of \approach\ by comparing AU accuracy under three different aggregation policies that aim to achieve single optimal policy from multiple objectives provided by different downstream AUs. We use four different (compatible but non-overlapping) AUs -- (1) Face-detection (FD) (2) Person-detection (PD), (3) Car-detection (CD) and (3) License-Plate Detection (LPD) in a enterprise building or bank scenario. We then evaluate \approach\ performance on a real-world deployment looking at the entrance of an enterprise building. 

%\input{policy}

%\input{local}

\subsection{AU-specific quality estimator}

For MORL to work well for multiple AUs running off of a single video
stream, it is very important for the rewards received for each of the
AUs to be highly accurate. These rewards for individual AUs are the
\textit{estimates} given by AU-specific quality estimator for each
AU. The higher the value of the \textit{estimate}, the better the chances
that the AU will give high accuracy on the frame. For evaluation of
the AU-specific quality estimator, we measure the Pearson and Spearman
correlation between the value of the \textit{estimate} and the actual AU
accuracy on the frame.

To evaluate AU-specific quality estimators for PD and FD, we use 2000
video frames from the video snippets extracted from
Roadway~\cite{canel2019scaling} and LSTN~\cite{fang2019locality}
datasets
%We first manually annotate the person and face bounding boxes using cvat-tool~\cite{cvat} 
and create 1 million variants by digitally transforming the original
frames using the python-pil library~\cite{pil}. These 1 million digitally
transformed video frames constitute the validation dataset for PD- and
FD-specific quality estimator's performance evaluation. To evaluate
CD-specific quality estimator, we create a validation dataset of 0.8
million digitally transformed images from the original 1.5K video
frames extracted from the Jackson and roadway
datasets~\cite{canel2019scaling}. For LPD-specific quality estimator
evaluation, we create validation dataset of 0.2 million digitally
transformed images from 500 original images extracted from
the Cars~\cite{krause20133d} and AOLP
datasets~\cite{hsu2012application}. Note that these validation datasets
are completely different from the training dataset used
in~\secref{subsec:au-quality}. Once the validation dataset is created,
we manually annotate the images using cvat-tool~\cite{cvat} and
measure the AU accuracy (mAP) on these frames using the annotation as
the ground truth~\footnote{We ensure that the ground-truth annotations
  are kept intact in the digitally transformed image variants as
  well.}. Next, we obtain the \textit{estimates} provided by the
AU-specific quality estimators on these images and then measure the
Pearson and Spearman correlation between the \textit{actual} accuracy
and the \textit{estimated} value provided by the AU-specific quality
estimators.

\begin{figure}
    \centering
        \includegraphics[width=0.4\textwidth]{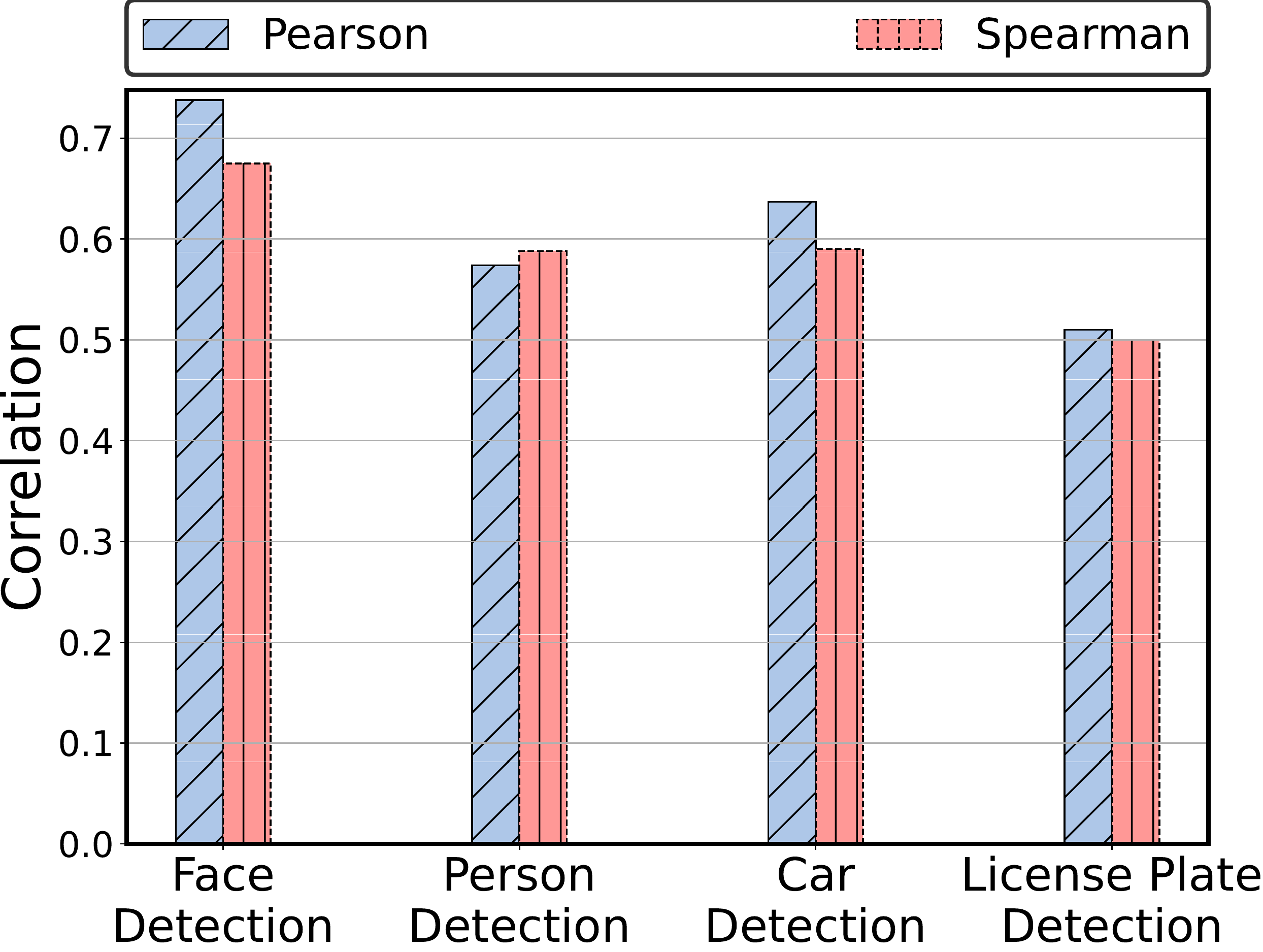}
        %\vspace{-0.1in}
        \caption{AU-specific quality estimator performance.}
        \label{fig:quality_estimator}
        \vspace{-0.1in}
\end{figure}

~\figref{fig:quality_estimator} shows a strong positive correlation
between the actual accuracy (mAP) and the provided quality estimate
for the four AUs (both Pearson and Spearman correlation are greater
than 0.5)~\cite{corr_1}. This shows that \textit{the newly developed
  AU-specific quality estimators provide reliable quality estimates
  that can be used by the MORL agent to effectively tune camera settings
  in order to improve accuracy for multiple AUs simultaneously.}

\label{subsec:quality-estimator}

\subsection{Aggregation strategy}
\label{subsec:aggregation_strategy}

\begin{figure}
    \centering
        \includegraphics[width=0.47\textwidth]{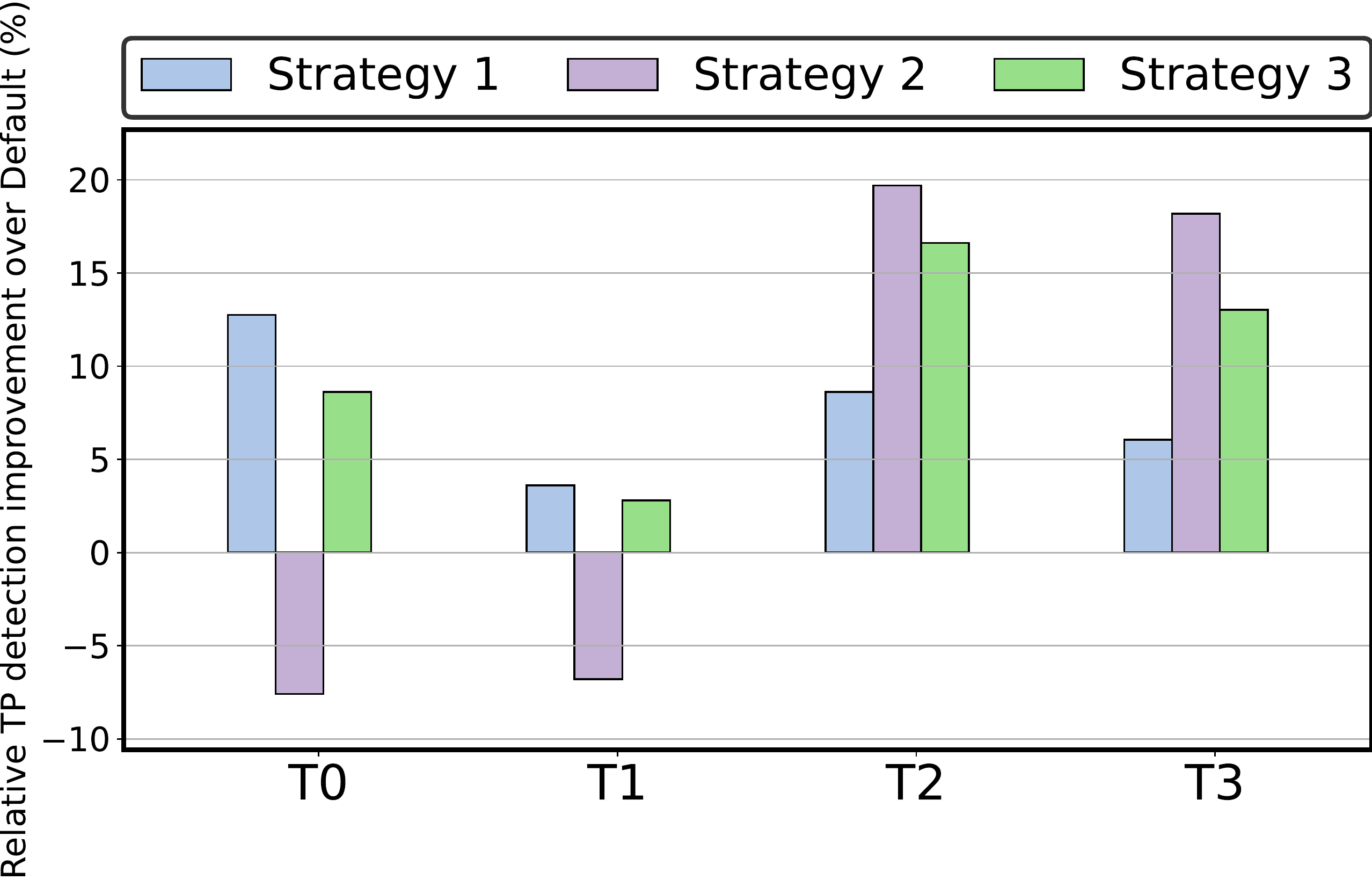}
        \vspace{-0.2in}
        \caption{Aggregation strategy performance.}
        %\caption{Impact of different MORL policies on cumulative true-positive detection count improvement over default setting considering four different analytics \ie FD, PD, CD, LPD (under controlled experimental setup).}
        \label{fig:strategy_perf}
        \vspace{-0.1in}
\end{figure}

Various aggregation strategies considered by the MORL agent in \approach\ is described in ~\secref{subsec:rl}. In this section, we compare these different strategies and present our experimental results. 

For comparing the three strategies, we use the original video captured
with default camera settings, in which two participants follow a
sequence of steps as described in
~\secref{subsec:optimal_setting_across_au}. We then annotate this
video and obtain the ground-truth on the original video. Next, we
create four variants of this video corresponding to different times in
the day (separated by at least 2 hours each) by using the 
%~\comment{virtual camera \ie that augments the input video with different environmental conditions through changing the post-capture digital transformation)} as
technique described in ~\cite{paul2021camtuner}, which applies digital transformation on the original video to generate a variant of the original video for a different time of the day, and then label these
videos as T0, T1, T2 and T3. These videos have 1000 frames each. Once
we have these four videos, we project them on a monitor screen in front of a real camera, as in ~\cite{paul2021camtuner}, and run four different AUs, i.e., PD, FD, CD
and LPD. We then compare the performance when we use the
three aggregation strategies, as described in ~\secref{subsec:rl}. For
Strategy 2 (weighted), we used predefined weights of 3, 2, 1 and 1
for FD, LPD, PD and CD, respectively, based on each AU's granularity,
i.e., finer the granularity of detection, more is the
weight. ~\figref{fig:strategy_perf} shows the performance of the three
aggregation strategies. We see that \textit{Strategy 3
  (winner-takes-all) performs the best when compared to the other two
  strategies, i.e., Strategy 1 (linear) and Strategy 2 (weighted).}

\subsection{Real-world deployment}
\label{subsec:real-world-expr}
\begin{figure}
    \centering
        \includegraphics[width=0.7\linewidth]{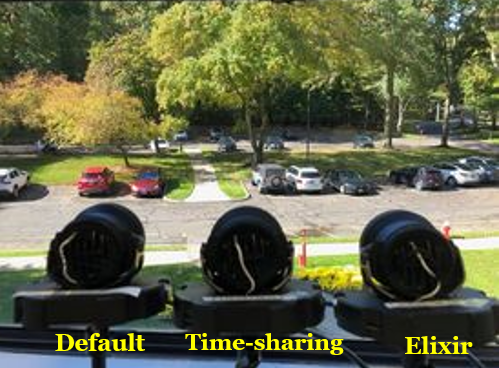}
        %\vspace{-0.1in}
        \caption{Real-world deployment setup.}
        \label{fig:real-world-deployment}
        \vspace{-0.1in}
\end{figure}

\begin{figure}[tb]
\begin{subfigure}[]{0.97\linewidth}
\centering
    \includegraphics[height=1.45in]{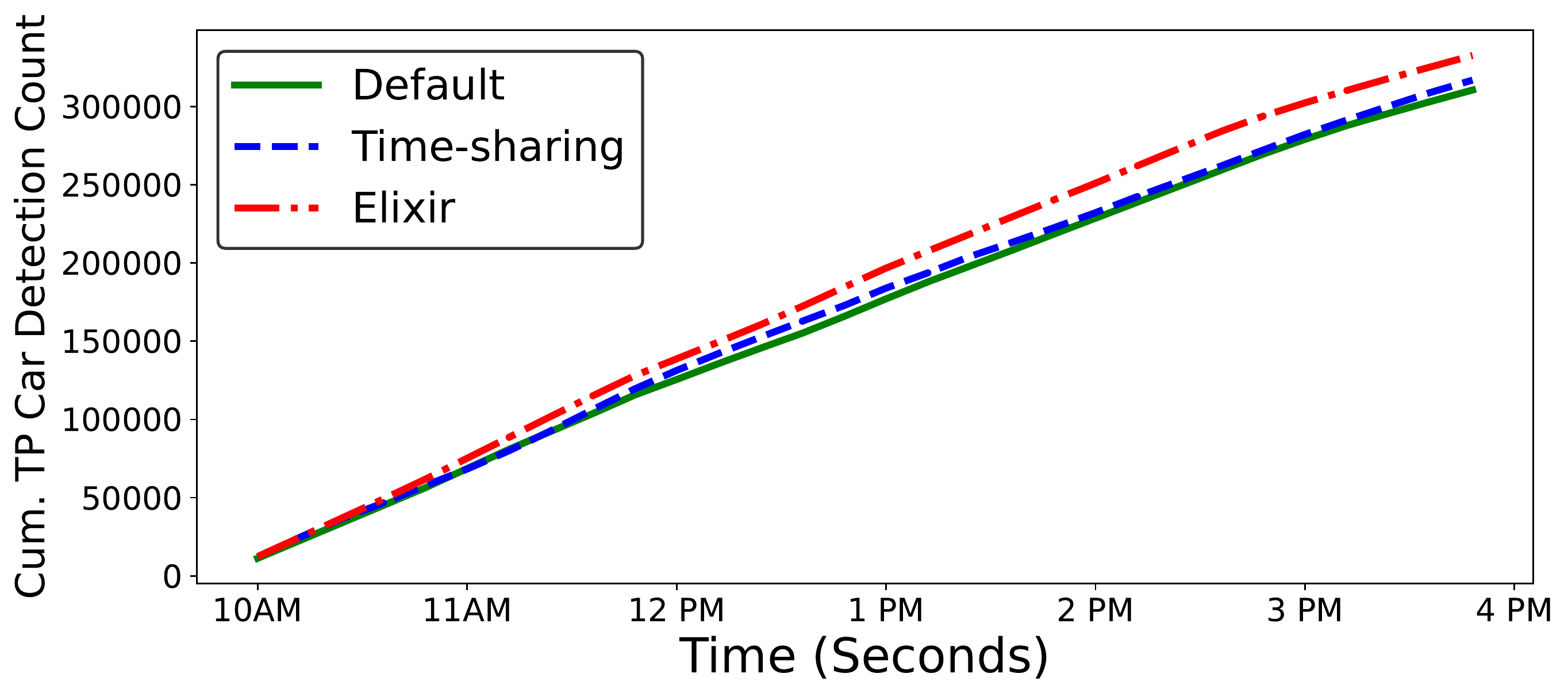}
    \vspace{-0.05in}
    \caption{Car Detection}
    \label{fig:daylong_car}
\end{subfigure}%
\hfill
\begin{subfigure}[]{0.995\linewidth}
\centering
    \includegraphics[height=1.5 in]{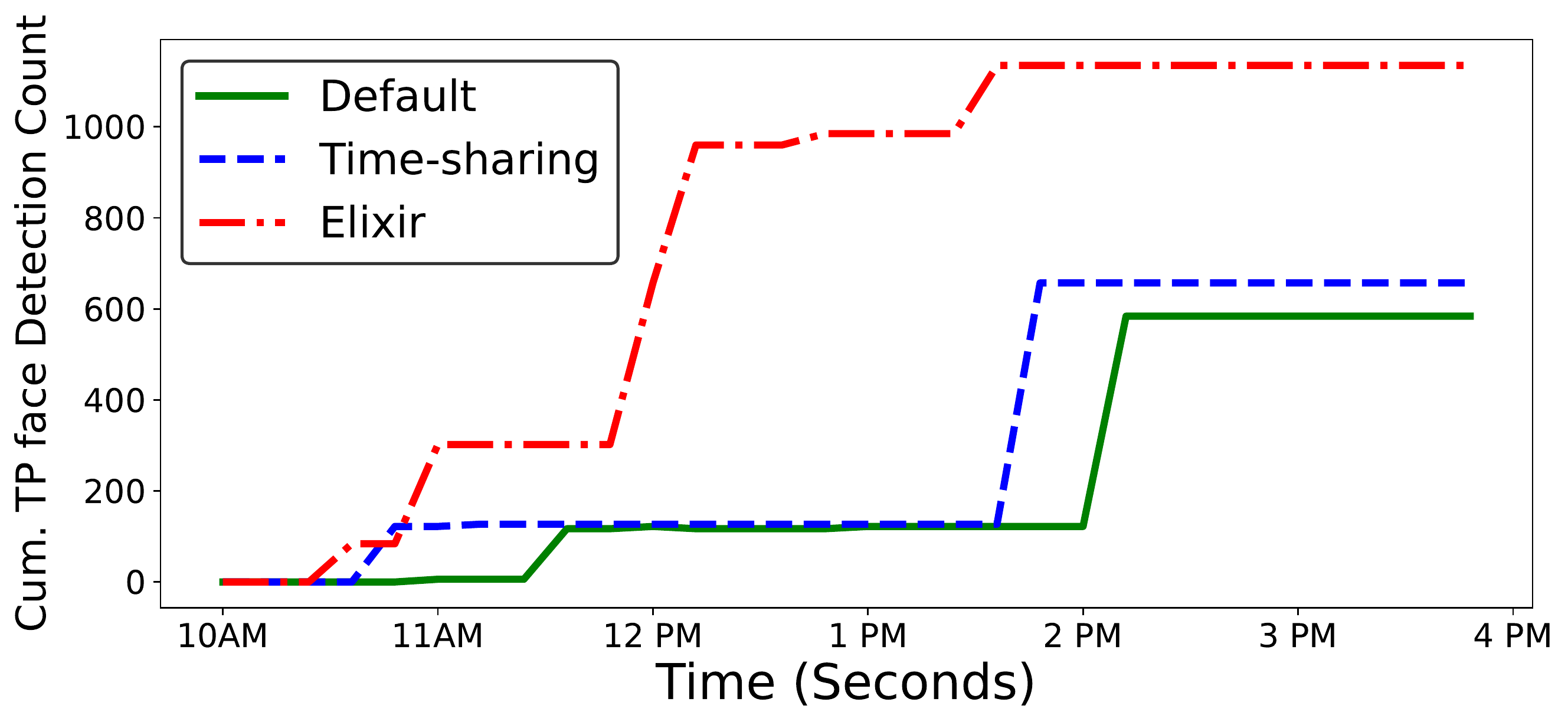}
    \vspace{-0.05in}
    \caption{Face Detection}
    \label{fig:daylong_face}
\end{subfigure}
\hfill
\begin{subfigure}[]{0.995\linewidth}
\centering
    \includegraphics[height=1.5 in]{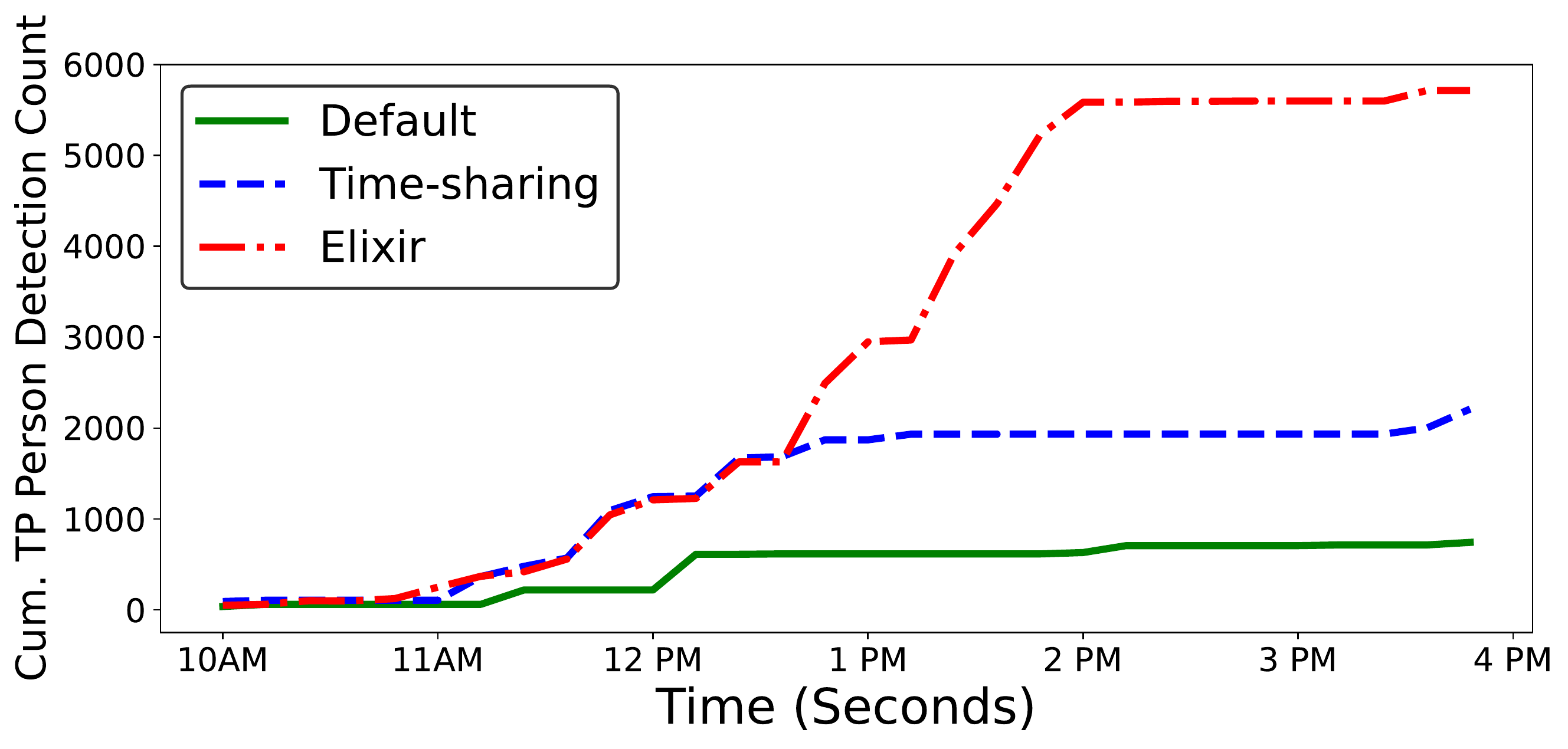}
    \vspace{-0.05in}
    \caption{Person Detection}
    \label{fig:daylong_person}
\end{subfigure}
\hfill
\begin{subfigure}[]{0.995\linewidth}
\centering
    \includegraphics[height=1.5 in]{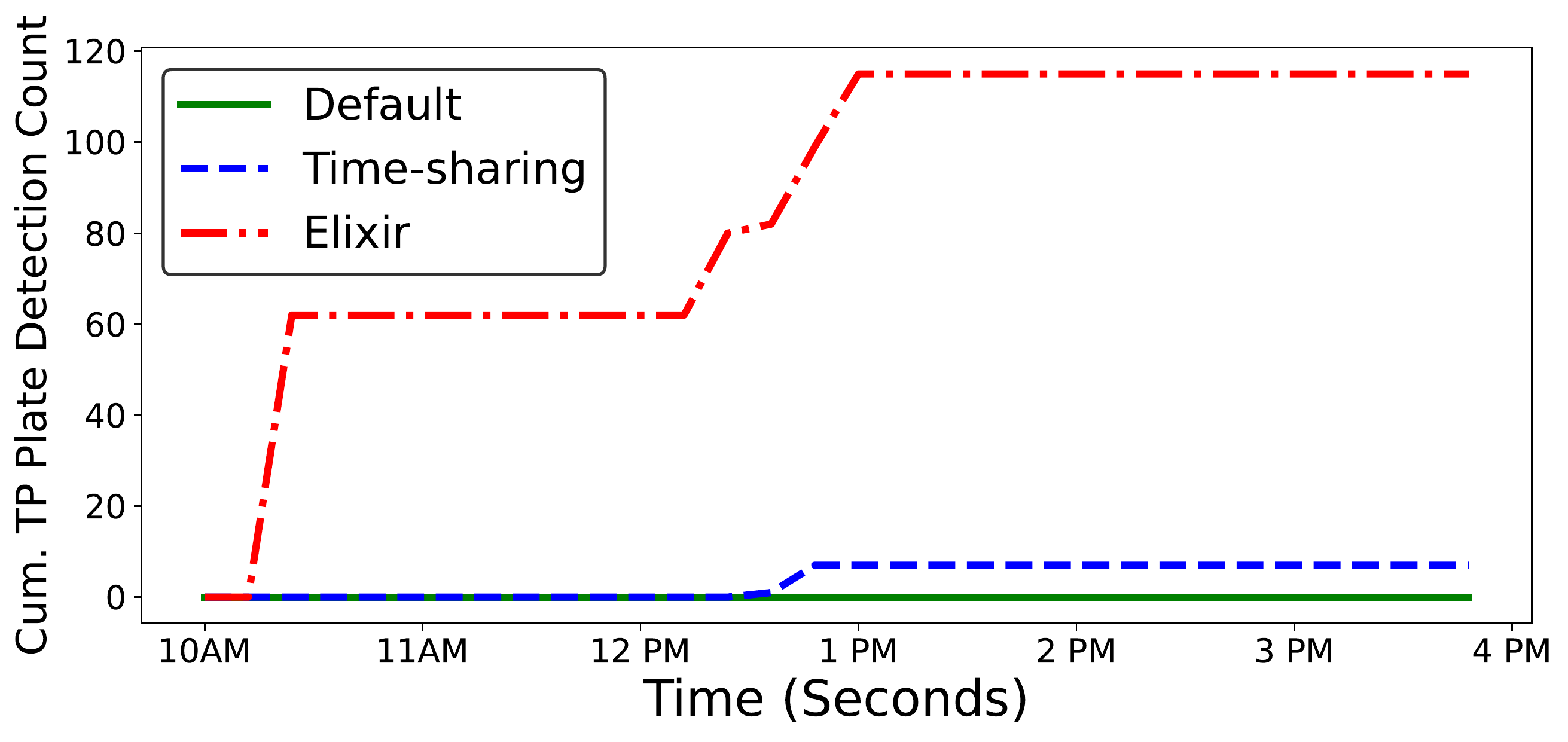}
    \vspace{-0.05in}
    \caption{License Plate Detection}
    \label{fig:daylong_plate}
\end{subfigure}
\vspace{-0.05in}
 \caption{\approach's performance comparison with alternative approaches when four AUs are running simultaneously.}
 \vspace{-0.3in}
  \label{fig:daylong_real}
\end{figure}

To evaluate performance of \approach\ in a real-world deployment, we
place three cameras next to each other as shown in
~\figref{fig:real-world-deployment}. All of them are identical
cameras, i.e., AXIS Q3515 Network cameras and they overlook a large
enterprise parking lot, where people park their cars and enter/exit
the building. With this deployment, we are able to run multiple AUs,
i.e., PD, FD, CD and LPD simultaneously on the video feed coming from a
single camera. The purpose of having three cameras is to compare
\approach\ with other two alternative approaches. The different
approaches being used on these three cameras is listed below:

\begin{enumerate}
    \item \textbf{Camera 1 (Default)}: Camera settings is always set to default camera settings and is never changed. All four AUs run simultaneously on the video feed from this camera with default camera settings.
    \item \textbf{Camera 2 (Time-sharing)}: Time-sharing technique is used on this camera, where each AU gets a fixed time-slot of 10 minutes and within this time-slot, camera settings are dynamically changed in favor of the specific AU for which the time-slot is allocated. Here, we deploy CamTuner ~\cite{paul2021camtuner} for each AU within its time-slot and time-slots are allocated in round-robin manner to each AU. Note that all four AUs are running simultaneously all the time, only the camera settings are changed periodically, favoring the AU for which the time-slot is allocated.
    \item \textbf{Camera 3 (\approach)}: We deploy \approach, which
      dynamically tunes camera settings to common-optimal setting that
      enhances data quality such that accuracy of all four AUs, i.e.,
      PD, FD, CD and LPD is improved simultaneously
\end{enumerate}

Each of these cameras upload the captured frames over 5G network to a remote edge-server (with 3.70 GHz Xeon(R) W-2145 CPU and an NVIDIA GeForce RTX 2080Ti GPU), where the four AUs are running. We use Yolov5~\cite{glenn_jocher_2022_6222936} object detection model for PD and CD~\footnote{Due to inter-class variability within same detector, we evaluate performance of person and car detection separately.}, Neofacev4~\cite{NIST-NEC} face detection model for FD and openalpr~\cite{openalpr} license plate detection model for LPD.
%running Yolov5~\cite{glenn_jocher_2022_6222936} object detection model (used for PD and CD~\footnote{Due to inter-class variability within same detector, we evaluate performance of person and car detection separately.}), Neofacev4 \textcolor{blue}{[add citation]} face detection model (used for FD) and openalpr \textcolor{blue}{[add citation]} license plate detection model (used for LPD). 
For Time-sharing and \approach\ approaches, the captured frames are also sent to a low-end Intel-NUC box (with 2.6 GHz Intel i7-6770HQ CPU) where the four AU-specific quality estimators run and provide quality \textit{estimates} on each frame for the four AUs, i.e., PD, FD, CD and LPD. 

%To populate the Q-tables corresponding to each AU, we perform online exploration for few hours and then start exploitation and dynamically adjust camera settings every 30 seconds.

We initially perform online exploration for few hours and then start exploitation to dynamically adjust camera settings. We adjust camera settings every 30 seconds. During exploitation, we run the four AUs (on each camera) for 6 continuous hours in a day and record the videos from each camera for manual inspection. 
%Upon manual inspection of the videos, we confirmed that \approach\ does not have any false positive detections for car, person, face or license plates. 
~\figref{fig:daylong_real} shows the cumulative number of true-positive car, face, person and license plate detections over the entire 6 hours for the three approaches~\footnote{Upon manual inspection of the videos, we confirmed that \approach\ does not have any false positive detections for car, face, person or license plates.}. ~\figref{fig:daylong_car} shows that \approach\ is able to detect 7.1\% (22,068) and 5.0\% (15,731) more cars than default and time-sharing approaches, respectively. ~\figref{fig:daylong_face} shows that 94\% (551) and 72\% (478) more faces are correctly being detected by \approach\ over the default and time-sharing approaches, respectively. \approach\ also detects 670.4\% (4975) and 158.6\% (3507) more persons than the default and time-sharing approaches, respectively, as shown in ~\figref{fig:daylong_person}. For license plate detection, we see that with default setting approach, none of the license plates were detected, with time-sharing approach, only 7 license plates were detected, while, with \approach, we are able to detect 115 license plates, as shown in ~\figref{fig:daylong_plate}. \textit{Thus, \approach\ is effective in dynamically adjusting camera settings to enhance quality of data such that accuracy of multiple AUs is improved simultaneously.}

\vspace{-0.05in}
\subsection{System Performance}
\label{subsec:overhead}

When \approach\ is deployed in operation, there are three key
additional tasks that occur, when compared to normal operation with
default camera settings. These three tasks are: (1) changing the
camera settings, (2) computing AU-specific quality estimation for all
the AUs, and (3) evaluating aggregation function to combine Q-values
from all the AUs. 
For (1), we use Axis' VAPIX library~\cite{vapix} to
programmatically change camera setting. This change in camera setting
takes $\sim$ 200 ms. Since different AU-specific quality estimators
use the same backbone and they run in parallel. For (2), it takes only
about 48 ms to compute all AU-specific quality estimates while the CPU
utilization is only around 45\%. Finally, for (3), the MORL agent
takes only about 1 ms to evaluate the aggregate function and update
the combined value. Since \approach\ runs in parallel with the AUs, it
does not incur any additional delay.  With respect to the delay over
5G network to upload the video frame, we observe that the latency to
upload a frame is only about 39.7 ms.

%The online operation of \approach\ involves three key tasks: changing the video camera parameters, AU-specific quality estimation for immediate reward to MORL agent, and evaluating the Q-functions for the agent. Changing the physical camera settings on AXIS Q3515 network camera using Restful APIs takes $\sim$ 200 ms. We run \approach\ on a low-end IoT device, an Intel-NUC box equipped with a 2.6 GHz Intel i7-6770HQ CPU. Since, different AU-specific quality estimators use the same backbone and runs in parallel, it takes 48 ms to get all AU-specific quality estimates while the CPU utilization is only around 45\%, and MORL agent takes around 1 ms to update all Q-tables. The captured frames are uploaded for multiple downstream analytics task on the remote edge-server which is equipped with Xeon processor and an NVIDIA GeForce RTX 2080Ti GPU over a 5G network with an average frame uploading latency of 39.7 ms. Since \approach\ runs in parallel with these AUs, it does not incur any additional delay to the pipeline. 

%\vspace{-0.05in}
%\section{Discussion: Further improvement over \approach}
%\section{Discussion}
%\label{sec:discussion}
\section{Further improvement over \approach}
\label{sec:further_improvement}

\begin{table}[tb]
\centering
\caption{Improvement over \approach}
\begin{tabular}{|c|c|c|}
    \hline
     Setting & PD accuracy & FD accuracy \\
     & mAP$_{PD}$(\%) & mAP$_{FD}$(\%)  \\
     & (DAY) & (NIGHT) \\
    \hline
    \emph{Common-optimal (\approach)}  & 91.7 & 48.8 \\
    \hline
    \emph{Common-optimal (\approach)}  & 98.8 & 50.0\\
    \emph{+ AU-specific } & & \\
    \emph{digital transformation} & & \\
    \hline
\end{tabular}
\label{tab:further_improvement}
 \vspace{-0.1in}
\end{table}
 
% In~\secref{subsec:optimal_settings_conflict}, as shown 
In~\tabref{tab:night-overall}, we saw that FD achieves 48.8\% accuracy
using the common-optimal setting, while the highest achieved accuracy
using FD-optimal setting is 50\%. Also, for PD, 
~\tabref{tab:day-overall} shows that the accuracy achieved using the common-optimal
setting is 91.7\% while the highest achieved accuracy using the PD-optimal
setting is 99.36\%. Thus, there is a gap in the highest possible
accuracy and the one achieved by the common-optimal setting. 
In this section, we show that it is possible to reduce this gap, i.e.,
boost the accuracy further, by applying AU-specific digital
transformation on the camera feed for each AU, after \approach\ tunes
the camera settings to the common-optimal setting. Note that the
accuracy for PD during NIGHT and the accuracy for FD during DAY, as
shown in ~\tabref{tab:overall-optimal}, achieve the highest possible
accuracy (83.3\% and 100\%, respectively) already in using the
common-optimal setting. Hence, we do not highlight them here.

For this study, on the frame captured with the ``common-optimal'' setting
discussed in~\secref{subsec:optimal_settings_conflict}, we
exhaustively apply different digital transformations by using the PIL
library~\cite{pil} and run these frames through
%the yolov5 object detection model for PD and retinaNet face detection model for FD
PD and FD AUs. We then choose the digital transformation that provides
the highest accuracy. ~\tabref{tab:further_improvement} shows that the
accuracy for PD is further boosted from 91.7\% to 98.8\% and the accuracy
for FD is further boosted from 48.8\% to 50\%. \textit{Thus, it is
  indeed possible to further improve the accuracy of the AUs by
  applying AU-specific digital transformations on the frame, after
  \approach\ tunes the camera settings to the common-optimal setting which
  works well for multiple AUs.} Note that in ~\cite{paul2021camtuner},
it is shown that directly applying digital transformation on frames
does not improve accuracy, if the frame capture itself is of ``poor"
quality. Therefore, we did not directly apply digital transformations
on the frames captured with default camera settings. Instead, we first
let \approach\ tune the camera settings to the common-optimal setting in
order to enhance data quality, and then apply digital transformation
on them. This ``common-optimal + AU-specific digital transformation'' approach
gives further improvement in accuracy over \approach.

%\input{common_and_digital_transformation}

%Introduction of local elixir and how to make local and global elixir tightly coupled one. 

%\vspace{-0.05in}
\section{Conclusion}
\label{sec:conclusion}

When multiple video analytics tasks run on a single video stream, the
optimal camera settings for each one of them is different, and the optimal
camera setting for one, degrades accuracy of the other. Identifying
the optimal camera setting that works well for multiple AUs and
automatically adjusting it based on the changes in the environmental
conditions and the video content is a very challenging task. In this
paper, we present \approach, which leverages Multi-Objective
Reinforcement Learning (MORL) to dynamically adjust the camera
settings such that multiple objectives (corresponding to different AUs
in operation) are considered simultaneously by the RL agent to decide
on the optimal camera settings that improves accuracy of multiple AUs
at the same time. Through real-world experiments with four AUs running
off of a single video stream, we show that \approach\ correctly
detects 7.1\% (22,068) and 5.0\% (15,731) more cars, 94\% (551) and
72\% (478) more faces, and 670.4\% (4975) and 158.6\% (3507) more
persons than the default-setting and time-sharing
approaches, respectively. It also detects 115 license plates, far more
than the time-sharing approach (7) and the default setting (0).

\bibliographystyle{ACM-Reference-Format}
\bibliography{reference}


\begin{thebibliography}{32}

%%% the Web address.

\ifx \showCODEN    \undefined \def \showCODEN     #1{\unskip}     \fi
\ifx \showDOI      \undefined \def \showDOI       #1{#1}\fi
\ifx \showISBNx    \undefined \def \showISBNx     #1{\unskip}     \fi
\ifx \showISBNxiii \undefined \def \showISBNxiii  #1{\unskip}     \fi
\ifx \showISSN     \undefined \def \showISSN      #1{\unskip}     \fi
\ifx \showLCCN     \undefined \def \showLCCN      #1{\unskip}     \fi
\ifx \shownote     \undefined \def \shownote      #1{#1}          \fi
\ifx \showarticletitle \undefined \def \showarticletitle #1{#1}   \fi
\ifx \showURL      \undefined \def \showURL       {\relax}        \fi
% invisible to TeX
\providecommand\bibfield[2]{#2}
\providecommand\bibinfo[2]{#2}
\providecommand\natexlab[1]{#1}
\providecommand\showeprint[2][]{arXiv:#2}

\bibitem[\protect\citeauthoryear{Analytics}{Analytics}{2022}]%
        {digital-market-research}
\bibfield{author}{\bibinfo{person}{Allied Analytics}.}
  \bibinfo{year}{2022}\natexlab{}.
\newblock \bibinfo{title}{Global Smart Sensor Market to Reach \$91.37 Billion
  by 2027}.
\newblock
\newblock
\urldef\tempurl%
\url{https://www.digitaljournal.com/pr/global-smart-sensor-market-to-reach-91-37-billion-by-2027}
\showURL{%
\tempurl}


\bibitem[\protect\citeauthoryear{Canel, Kim, Zhou, Li, Lim, Andersen, Kaminsky,
  and Dulloor}{Canel et~al\mbox{.}}{2019}]%
        {canel2019scaling}
\bibfield{author}{\bibinfo{person}{Christopher Canel}, \bibinfo{person}{Thomas
  Kim}, \bibinfo{person}{Giulio Zhou}, \bibinfo{person}{Conglong Li},
  \bibinfo{person}{Hyeontaek Lim}, \bibinfo{person}{David~G Andersen},
  \bibinfo{person}{Michael Kaminsky}, {and} \bibinfo{person}{Subramanya
  Dulloor}.} \bibinfo{year}{2019}\natexlab{}.
\newblock \showarticletitle{Scaling video analytics on constrained edge nodes}.
\newblock \bibinfo{journal}{\emph{Proceedings of Machine Learning and Systems}}
   \bibinfo{volume}{1} (\bibinfo{year}{2019}), \bibinfo{pages}{406--417}.
\newblock


\bibitem[\protect\citeauthoryear{Clark and Contributors}{Clark and
  Contributors}{[n.d.]}]%
        {pil}
\bibfield{author}{\bibinfo{person}{Alex Clark} {and}
  \bibinfo{person}{Contributors}.} \bibinfo{year}{[n.d.]}\natexlab{}.
\newblock \bibinfo{title}{Pillow Library}.
\newblock
  \bibinfo{howpublished}{\url{https://pillow.readthedocs.io/en/stable/}}.
\newblock


\bibitem[\protect\citeauthoryear{Communications}{Communications}{[n.d.]}]%
        {vapix}
\bibfield{author}{\bibinfo{person}{Axis Communications}.}
  \bibinfo{year}{[n.d.]}\natexlab{}.
\newblock \bibinfo{title}{VAPIX Library}.
\newblock
\newblock
\urldef\tempurl%
\url{https://www.axis.com/vapix-library/}
\showURL{%
\tempurl}


\bibitem[\protect\citeauthoryear{Dao, Roy-Chowdhury, Nasrabadi, Krishnamurthy,
  Mohapatra, and Kaplan}{Dao et~al\mbox{.}}{2017}]%
        {dao2017accurate}
\bibfield{author}{\bibinfo{person}{Tuan Dao}, \bibinfo{person}{Amit
  Roy-Chowdhury}, \bibinfo{person}{Nasser Nasrabadi},
  \bibinfo{person}{Srikanth~V Krishnamurthy}, \bibinfo{person}{Prasant
  Mohapatra}, {and} \bibinfo{person}{Lance~M Kaplan}.}
  \bibinfo{year}{2017}\natexlab{}.
\newblock \showarticletitle{Accurate and timely situation awareness retrieval
  from a bandwidth constrained camera network}. In
  \bibinfo{booktitle}{\emph{2017 IEEE 14th International Conference on Mobile
  Ad Hoc and Sensor Systems (MASS)}}. IEEE, \bibinfo{pages}{416--425}.
\newblock


\bibitem[\protect\citeauthoryear{Fang, Zhan, Cai, Gao, and Hu}{Fang
  et~al\mbox{.}}{2019}]%
        {fang2019locality}
\bibfield{author}{\bibinfo{person}{Yanyan Fang}, \bibinfo{person}{Biyun Zhan},
  \bibinfo{person}{Wandi Cai}, \bibinfo{person}{Shenghua Gao}, {and}
  \bibinfo{person}{Bo Hu}.} \bibinfo{year}{2019}\natexlab{}.
\newblock \showarticletitle{Locality-constrained spatial transformer network
  for video crowd counting}. In \bibinfo{booktitle}{\emph{2019 IEEE
  international conference on multimedia and expo (ICME)}}. IEEE,
  \bibinfo{pages}{814--819}.
\newblock


\bibitem[\protect\citeauthoryear{Hayes, R{\u a}dulescu, Bargiacchi,
  K{\"a}llstr{\"o}m, Macfarlane, Reymond, Verstraeten, Zintgraf, Dazeley,
  Heintz, Howley, Irissappane, Mannion, Now{\'e}, Ramos, Restelli, Vamplew, and
  Roijers}{Hayes et~al\mbox{.}}{2022}]%
        {morl-2}
\bibfield{author}{\bibinfo{person}{Conor~F. Hayes}, \bibinfo{person}{Roxana
  R{\u a}dulescu}, \bibinfo{person}{Eugenio Bargiacchi}, \bibinfo{person}{Johan
  K{\"a}llstr{\"o}m}, \bibinfo{person}{Matthew Macfarlane},
  \bibinfo{person}{Mathieu Reymond}, \bibinfo{person}{Timothy Verstraeten},
  \bibinfo{person}{Luisa~M. Zintgraf}, \bibinfo{person}{Richard Dazeley},
  \bibinfo{person}{Fredrik Heintz}, \bibinfo{person}{Enda Howley},
  \bibinfo{person}{Athirai~A. Irissappane}, \bibinfo{person}{Patrick Mannion},
  \bibinfo{person}{Ann Now{\'e}}, \bibinfo{person}{Gabriel Ramos},
  \bibinfo{person}{Marcello Restelli}, \bibinfo{person}{Peter Vamplew}, {and}
  \bibinfo{person}{Diederik~M. Roijers}.} \bibinfo{year}{2022}\natexlab{}.
\newblock \showarticletitle{A practical guide to multi-objective reinforcement
  learning and planning}.
\newblock \bibinfo{journal}{\emph{Autonomous Agents and Multi-Agent Systems}}
  \bibinfo{volume}{36}, \bibinfo{number}{1} (\bibinfo{year}{2022}),
  \bibinfo{pages}{26}.
\newblock
\showISBNx{1573-7454}
\urldef\tempurl%
\url{https://doi.org/10.1007/s10458-022-09552-y}
\showDOI{\tempurl}


\bibitem[\protect\citeauthoryear{Hsieh, Ananthanarayanan, Bodik, Venkataraman,
  Bahl, Philipose, Gibbons, and Mutlu}{Hsieh et~al\mbox{.}}{2018}]%
        {222587}
\bibfield{author}{\bibinfo{person}{Kevin Hsieh}, \bibinfo{person}{Ganesh
  Ananthanarayanan}, \bibinfo{person}{Peter Bodik}, \bibinfo{person}{Shivaram
  Venkataraman}, \bibinfo{person}{Paramvir Bahl}, \bibinfo{person}{Matthai
  Philipose}, \bibinfo{person}{Phillip~B. Gibbons}, {and} \bibinfo{person}{Onur
  Mutlu}.} \bibinfo{year}{2018}\natexlab{}.
\newblock \showarticletitle{Focus: Querying Large Video Datasets with Low
  Latency and Low Cost}. In \bibinfo{booktitle}{\emph{13th {USENIX} Symposium
  on Operating Systems Design and Implementation ({OSDI} 18)}}.
  \bibinfo{publisher}{{USENIX} Association}, \bibinfo{address}{Carlsbad, CA},
  \bibinfo{pages}{269--286}.
\newblock
\showISBNx{978-1-939133-08-3}
\urldef\tempurl%
\url{https://www.usenix.org/conference/osdi18/presentation/hsieh}
\showURL{%
\tempurl}


\bibitem[\protect\citeauthoryear{Hsu, Chen, and Chung}{Hsu
  et~al\mbox{.}}{2012}]%
        {hsu2012application}
\bibfield{author}{\bibinfo{person}{Gee-Sern Hsu}, \bibinfo{person}{Jiun-Chang
  Chen}, {and} \bibinfo{person}{Yu-Zu Chung}.} \bibinfo{year}{2012}\natexlab{}.
\newblock \showarticletitle{Application-oriented license plate recognition}.
\newblock \bibinfo{journal}{\emph{IEEE transactions on vehicular technology}}
  \bibinfo{volume}{62}, \bibinfo{number}{2} (\bibinfo{year}{2012}),
  \bibinfo{pages}{552--561}.
\newblock


\bibitem[\protect\citeauthoryear{Jain, Nguyen, Gruteser, and Bahl}{Jain
  et~al\mbox{.}}{2017}]%
        {jain2017panoptes}
\bibfield{author}{\bibinfo{person}{Shubham Jain}, \bibinfo{person}{Viet
  Nguyen}, \bibinfo{person}{Marco Gruteser}, {and} \bibinfo{person}{Paramvir
  Bahl}.} \bibinfo{year}{2017}\natexlab{}.
\newblock \showarticletitle{Panoptes: Servicing multiple applications
  simultaneously using steerable cameras}. In
  \bibinfo{booktitle}{\emph{Proceedings of the 16th ACM/IEEE International
  Conference on Information Processing in Sensor Networks}}.
  \bibinfo{pages}{119--130}.
\newblock


\bibitem[\protect\citeauthoryear{Jang, Lee, Shin, and Lee}{Jang
  et~al\mbox{.}}{2018}]%
        {jang2018application}
\bibfield{author}{\bibinfo{person}{Si~Young Jang}, \bibinfo{person}{Yoonhyung
  Lee}, \bibinfo{person}{Byoungheon Shin}, {and} \bibinfo{person}{Dongman
  Lee}.} \bibinfo{year}{2018}\natexlab{}.
\newblock \showarticletitle{Application-aware IoT camera virtualization for
  video analytics edge computing}. In \bibinfo{booktitle}{\emph{2018 IEEE/ACM
  Symposium on Edge Computing (SEC)}}. IEEE, \bibinfo{pages}{132--144}.
\newblock


\bibitem[\protect\citeauthoryear{Jiang, Ananthanarayanan, Bodik, Sen, and
  Stoica}{Jiang et~al\mbox{.}}{2018}]%
        {jiang2018chameleon}
\bibfield{author}{\bibinfo{person}{Junchen Jiang}, \bibinfo{person}{Ganesh
  Ananthanarayanan}, \bibinfo{person}{Peter Bodik}, \bibinfo{person}{Siddhartha
  Sen}, {and} \bibinfo{person}{Ion Stoica}.} \bibinfo{year}{2018}\natexlab{}.
\newblock \showarticletitle{Chameleon: scalable adaptation of video analytics}.
  In \bibinfo{booktitle}{\emph{Proceedings of the 2018 Conference of the ACM
  Special Interest Group on Data Communication}}. \bibinfo{pages}{253--266}.
\newblock


\bibitem[\protect\citeauthoryear{Jocher, Chaurasia, Stoken, Borovec,
  NanoCode012, Kwon, TaoXie, Fang, imyhxy, Michael, Lorna, V, Montes, Nadar,
  Laughing, tkianai, yxNONG, Skalski, Wang, Hogan, Fati, Mammana, AlexWang1900,
  Patel, Yiwei, You, Hajek, Diaconu, and Minh}{Jocher et~al\mbox{.}}{2022}]%
        {glenn_jocher_2022_6222936}
\bibfield{author}{\bibinfo{person}{Glenn Jocher}, \bibinfo{person}{Ayush
  Chaurasia}, \bibinfo{person}{Alex Stoken}, \bibinfo{person}{Jirka Borovec},
  \bibinfo{person}{NanoCode012}, \bibinfo{person}{Yonghye Kwon},
  \bibinfo{person}{TaoXie}, \bibinfo{person}{Jiacong Fang},
  \bibinfo{person}{imyhxy}, \bibinfo{person}{Kalen Michael},
  \bibinfo{person}{Lorna}, \bibinfo{person}{Abhiram V}, \bibinfo{person}{Diego
  Montes}, \bibinfo{person}{Jebastin Nadar}, \bibinfo{person}{Laughing},
  \bibinfo{person}{tkianai}, \bibinfo{person}{yxNONG}, \bibinfo{person}{Piotr
  Skalski}, \bibinfo{person}{Zhiqiang Wang}, \bibinfo{person}{Adam Hogan},
  \bibinfo{person}{Cristi Fati}, \bibinfo{person}{Lorenzo Mammana},
  \bibinfo{person}{AlexWang1900}, \bibinfo{person}{Deep Patel},
  \bibinfo{person}{Ding Yiwei}, \bibinfo{person}{Felix You},
  \bibinfo{person}{Jan Hajek}, \bibinfo{person}{Laurentiu Diaconu}, {and}
  \bibinfo{person}{Mai~Thanh Minh}.} \bibinfo{year}{2022}\natexlab{}.
\newblock \bibinfo{booktitle}{\emph{{ultralytics/yolov5: v6.1 - TensorRT,
  TensorFlow Edge TPU and OpenVINO Export and Inference}}}.
\newblock
\urldef\tempurl%
\url{https://doi.org/10.5281/zenodo.6222936}
\showDOI{\tempurl}


\bibitem[\protect\citeauthoryear{Kang, Emmons, Abuzaid, Bailis, and
  Zaharia}{Kang et~al\mbox{.}}{2017}]%
        {10.14778/3137628.3137664}
\bibfield{author}{\bibinfo{person}{Daniel Kang}, \bibinfo{person}{John Emmons},
  \bibinfo{person}{Firas Abuzaid}, \bibinfo{person}{Peter Bailis}, {and}
  \bibinfo{person}{Matei Zaharia}.} \bibinfo{year}{2017}\natexlab{}.
\newblock \showarticletitle{NoScope: Optimizing Neural Network Queries over
  Video at Scale}.
\newblock \bibinfo{journal}{\emph{Proc. VLDB Endow.}} \bibinfo{volume}{10},
  \bibinfo{number}{11} (\bibinfo{date}{Aug.} \bibinfo{year}{2017}),
  \bibinfo{pages}{1586–1597}.
\newblock
\showISSN{2150-8097}
\urldef\tempurl%
\url{https://doi.org/10.14778/3137628.3137664}
\showDOI{\tempurl}


\bibitem[\protect\citeauthoryear{Khani, Hamadanian, Nasr-Esfahany, and
  Alizadeh}{Khani et~al\mbox{.}}{2021}]%
        {khani2021real}
\bibfield{author}{\bibinfo{person}{Mehrdad Khani}, \bibinfo{person}{Pouya
  Hamadanian}, \bibinfo{person}{Arash Nasr-Esfahany}, {and}
  \bibinfo{person}{Mohammad Alizadeh}.} \bibinfo{year}{2021}\natexlab{}.
\newblock \showarticletitle{Real-time video inference on edge devices via
  adaptive model streaming}. In \bibinfo{booktitle}{\emph{Proceedings of the
  IEEE/CVF International Conference on Computer Vision}}.
  \bibinfo{pages}{4572--4582}.
\newblock


\bibitem[\protect\citeauthoryear{Kingma and Ba}{Kingma and Ba}{2014}]%
        {kingma2014adam}
\bibfield{author}{\bibinfo{person}{Diederik~P Kingma} {and}
  \bibinfo{person}{Jimmy Ba}.} \bibinfo{year}{2014}\natexlab{}.
\newblock \showarticletitle{Adam: A method for stochastic optimization}.
\newblock \bibinfo{journal}{\emph{arXiv preprint arXiv:1412.6980}}
  (\bibinfo{year}{2014}).
\newblock


\bibitem[\protect\citeauthoryear{Krause, Stark, Deng, and Fei-Fei}{Krause
  et~al\mbox{.}}{2013}]%
        {krause20133d}
\bibfield{author}{\bibinfo{person}{Jonathan Krause}, \bibinfo{person}{Michael
  Stark}, \bibinfo{person}{Jia Deng}, {and} \bibinfo{person}{Li Fei-Fei}.}
  \bibinfo{year}{2013}\natexlab{}.
\newblock \showarticletitle{3d object representations for fine-grained
  categorization}. In \bibinfo{booktitle}{\emph{Proceedings of the IEEE
  international conference on computer vision workshops}}.
  \bibinfo{pages}{554--561}.
\newblock


\bibitem[\protect\citeauthoryear{Lin, Maire, Belongie, Hays, Perona, Ramanan,
  Doll{\'a}r, and Zitnick}{Lin et~al\mbox{.}}{2014}]%
        {lin2014microsoft}
\bibfield{author}{\bibinfo{person}{Tsung-Yi Lin}, \bibinfo{person}{Michael
  Maire}, \bibinfo{person}{Serge Belongie}, \bibinfo{person}{James Hays},
  \bibinfo{person}{Pietro Perona}, \bibinfo{person}{Deva Ramanan},
  \bibinfo{person}{Piotr Doll{\'a}r}, {and} \bibinfo{person}{C~Lawrence
  Zitnick}.} \bibinfo{year}{2014}\natexlab{}.
\newblock \showarticletitle{Microsoft coco: Common objects in context}. In
  \bibinfo{booktitle}{\emph{European conference on computer vision}}. Springer,
  \bibinfo{pages}{740--755}.
\newblock


\bibitem[\protect\citeauthoryear{Liu, Xu, and Hu}{Liu et~al\mbox{.}}{2014}]%
        {liu2014multiobjective}
\bibfield{author}{\bibinfo{person}{Chunming Liu}, \bibinfo{person}{Xin Xu},
  {and} \bibinfo{person}{Dewen Hu}.} \bibinfo{year}{2014}\natexlab{}.
\newblock \showarticletitle{Multiobjective reinforcement learning: A
  comprehensive overview}.
\newblock \bibinfo{journal}{\emph{IEEE Transactions on Systems, Man, and
  Cybernetics: Systems}} \bibinfo{volume}{45}, \bibinfo{number}{3}
  (\bibinfo{year}{2014}), \bibinfo{pages}{385--398}.
\newblock


\bibitem[\protect\citeauthoryear{NEC}{NEC}{2021}]%
        {NIST-NEC}
\bibfield{author}{\bibinfo{person}{NEC}.} \bibinfo{year}{2021}\natexlab{}.
\newblock \bibinfo{title}{NEC Face Recognition Technology Ranks First in NIST
  Accuracy Testing}.
\newblock
  \bibinfo{howpublished}{\url{https://www.nec.com/en/press/202108/global_20210823_01.html}}.
\newblock


\bibitem[\protect\citeauthoryear{openalpr}{openalpr}{[n.d.]}]%
        {openalpr}
\bibfield{author}{\bibinfo{person}{openalpr}.}
  \bibinfo{year}{[n.d.]}\natexlab{}.
\newblock \bibinfo{title}{Open-source automatic license plate recognition}.
\newblock
\newblock
\urldef\tempurl%
\url{https://github.com/openalpr/openalpr}
\showURL{%
\tempurl}


\bibitem[\protect\citeauthoryear{openvinotoolkit}{openvinotoolkit}{[n.d.]}]%
        {cvat}
\bibfield{author}{\bibinfo{person}{openvinotoolkit}.}
  \bibinfo{year}{[n.d.]}\natexlab{}.
\newblock \bibinfo{title}{Computer Vision Annotation Tool (CVAT)}.
\newblock
  \bibinfo{howpublished}{\url{https://github.com/openvinotoolkit/cvat}}.
\newblock


\bibitem[\protect\citeauthoryear{Padmanabhan, Iyer, Ananthanarayanan, Shu,
  Karianakis, Xu, and Netravali}{Padmanabhan et~al\mbox{.}}{[n.d.]}]%
        {padmanabhantowards}
\bibfield{author}{\bibinfo{person}{Arthi Padmanabhan},
  \bibinfo{person}{Anand~Padmanabha Iyer}, \bibinfo{person}{Ganesh
  Ananthanarayanan}, \bibinfo{person}{Yuanchao Shu}, \bibinfo{person}{Nikolaos
  Karianakis}, \bibinfo{person}{Guoqing~Harry Xu}, {and} \bibinfo{person}{Ravi
  Netravali}.} \bibinfo{year}{[n.d.]}\natexlab{}.
\newblock \showarticletitle{Towards {M}emory-{E}fficient {I}nference in {E}dge
  {V}ideo {A}nalytics}.
\newblock  (\bibinfo{year}{[n.\,d.]}).
\newblock


\bibitem[\protect\citeauthoryear{Paul, Rao, Coviello, Sankaradas, Po, Hu, and
  Chakradhar}{Paul et~al\mbox{.}}{2021}]%
        {paul2021camtuner}
\bibfield{author}{\bibinfo{person}{Sibendu Paul}, \bibinfo{person}{Kunal Rao},
  \bibinfo{person}{Giuseppe Coviello}, \bibinfo{person}{Murugan Sankaradas},
  \bibinfo{person}{Oliver Po}, \bibinfo{person}{Y~Charlie Hu}, {and}
  \bibinfo{person}{Srimat~T Chakradhar}.} \bibinfo{year}{2021}\natexlab{}.
\newblock \showarticletitle{CamTuner: Reinforcement-Learning based System for
  Camera Parameter Tuning to enhance Analytics}.
\newblock \bibinfo{journal}{\emph{arXiv preprint arXiv:2107.03964}}
  (\bibinfo{year}{2021}).
\newblock


\bibitem[\protect\citeauthoryear{Qureshi and Terzopoulos}{Qureshi and
  Terzopoulos}{2009}]%
        {qureshi2009planning}
\bibfield{author}{\bibinfo{person}{Faisal~Z Qureshi} {and}
  \bibinfo{person}{Demetri Terzopoulos}.} \bibinfo{year}{2009}\natexlab{}.
\newblock \showarticletitle{Planning ahead for PTZ camera assignment and
  handoff}. In \bibinfo{booktitle}{\emph{2009 Third ACM/IEEE International
  Conference on Distributed Smart Cameras (ICDSC)}}. IEEE,
  \bibinfo{pages}{1--8}.
\newblock


\bibitem[\protect\citeauthoryear{Sharma, Irwin, Shenoy, and Zink}{Sharma
  et~al\mbox{.}}{2011}]%
        {sharma2011multisense}
\bibfield{author}{\bibinfo{person}{Navin~K Sharma}, \bibinfo{person}{David~E
  Irwin}, \bibinfo{person}{Prashant~J Shenoy}, {and} \bibinfo{person}{Michael
  Zink}.} \bibinfo{year}{2011}\natexlab{}.
\newblock \showarticletitle{Multisense: fine-grained multiplexing for steerable
  camera sensor networks}. In \bibinfo{booktitle}{\emph{Proceedings of the
  second annual ACM conference on Multimedia systems}}.
  \bibinfo{pages}{23--34}.
\newblock


\bibitem[\protect\citeauthoryear{Szegedy, Vanhoucke, Ioffe, Shlens, and
  Wojna}{Szegedy et~al\mbox{.}}{2016}]%
        {inception-v3}
\bibfield{author}{\bibinfo{person}{Christian Szegedy}, \bibinfo{person}{Vincent
  Vanhoucke}, \bibinfo{person}{Sergey Ioffe}, \bibinfo{person}{Jon Shlens},
  {and} \bibinfo{person}{Zbigniew Wojna}.} \bibinfo{year}{2016}\natexlab{}.
\newblock \showarticletitle{Rethinking the {I}nception {A}rchitecture for
  {C}omputer {V}ision}. In \bibinfo{booktitle}{\emph{2016 IEEE Conference on
  Computer Vision and Pattern Recognition (CVPR)}}.
  \bibinfo{pages}{2818--2826}.
\newblock
\urldef\tempurl%
\url{https://doi.org/10.1109/CVPR.2016.308}
\showDOI{\tempurl}


\bibitem[\protect\citeauthoryear{thebmj}{thebmj}{2019}]%
        {corr_1}
\bibfield{author}{\bibinfo{person}{thebmj}.} \bibinfo{year}{2019}\natexlab{}.
\newblock \bibinfo{title}{correlation-and-regression}.
\newblock
  \bibinfo{howpublished}{\href{https://www.bmj.com/about-bmj/resources-readers/publications/statistics-square-one/11-correlation-and-regression}{correlation-range}}.
\newblock


\bibitem[\protect\citeauthoryear{Tong, Li, Li, Huang, and Hua}{Tong
  et~al\mbox{.}}{2021}]%
        {tong2021large}
\bibfield{author}{\bibinfo{person}{Panrong Tong}, \bibinfo{person}{Mingqian
  Li}, \bibinfo{person}{Mo Li}, \bibinfo{person}{Jianqiang Huang}, {and}
  \bibinfo{person}{Xiansheng Hua}.} \bibinfo{year}{2021}\natexlab{}.
\newblock \showarticletitle{Large-scale vehicle trajectory reconstruction with
  camera sensing network}. In \bibinfo{booktitle}{\emph{Proceedings of the 27th
  Annual International Conference on Mobile Computing and Networking}}.
  \bibinfo{pages}{188--200}.
\newblock


\bibitem[\protect\citeauthoryear{Yao, Chen, Koschan, and Abidi}{Yao
  et~al\mbox{.}}{2010}]%
        {yao2010adaptive}
\bibfield{author}{\bibinfo{person}{Yi Yao}, \bibinfo{person}{Chung-Hao Chen},
  \bibinfo{person}{Andreas Koschan}, {and} \bibinfo{person}{Mongi Abidi}.}
  \bibinfo{year}{2010}\natexlab{}.
\newblock \showarticletitle{Adaptive online camera coordination for
  multi-camera multi-target surveillance}.
\newblock \bibinfo{journal}{\emph{Computer Vision and Image Understanding}}
  \bibinfo{volume}{114}, \bibinfo{number}{4} (\bibinfo{year}{2010}),
  \bibinfo{pages}{463--474}.
\newblock


\bibitem[\protect\citeauthoryear{Zhang, Jin, Ratnasamy, Wawrzynek, and
  Lee}{Zhang et~al\mbox{.}}{2018}]%
        {zhang2018awstream}
\bibfield{author}{\bibinfo{person}{Ben Zhang}, \bibinfo{person}{Xin Jin},
  \bibinfo{person}{Sylvia Ratnasamy}, \bibinfo{person}{John Wawrzynek}, {and}
  \bibinfo{person}{Edward~A Lee}.} \bibinfo{year}{2018}\natexlab{}.
\newblock \showarticletitle{Awstream: Adaptive wide-area streaming analytics}.
  In \bibinfo{booktitle}{\emph{Proceedings of the 2018 Conference of the ACM
  Special Interest Group on Data Communication}}. \bibinfo{pages}{236--252}.
\newblock


\bibitem[\protect\citeauthoryear{Zhang, Ananthanarayanan, Bodik, Philipose,
  Bahl, and Freedman}{Zhang et~al\mbox{.}}{2017}]%
        {201465}
\bibfield{author}{\bibinfo{person}{Haoyu Zhang}, \bibinfo{person}{Ganesh
  Ananthanarayanan}, \bibinfo{person}{Peter Bodik}, \bibinfo{person}{Matthai
  Philipose}, \bibinfo{person}{Paramvir Bahl}, {and}
  \bibinfo{person}{Michael~J. Freedman}.} \bibinfo{year}{2017}\natexlab{}.
\newblock \showarticletitle{Live Video Analytics at Scale with Approximation
  and Delay-Tolerance}. In \bibinfo{booktitle}{\emph{14th {USENIX} Symposium on
  Networked Systems Design and Implementation ({NSDI} 17)}}.
  \bibinfo{publisher}{{USENIX} Association}, \bibinfo{address}{Boston, MA},
  \bibinfo{pages}{377--392}.
\newblock
\showISBNx{978-1-931971-37-9}


\end{thebibliography}

\end{document}